\documentclass[pdflatex,sn-mathphys-num]{sn-jnl}

\usepackage{graphicx}
\usepackage{amsmath,amssymb}
\usepackage{booktabs}
\usepackage{float}
\usepackage{algorithm}
\usepackage{algorithmic}
\usepackage{multirow}
\usepackage{array}
\usepackage{longtable}
\usepackage{url}
\usepackage{listings}
\usepackage{xcolor}
\usepackage{tabularx}
\usepackage{microtype}
\usepackage{placeins}
\usepackage{enumitem}
\usepackage{wrapfig}

\title[Graph Neural Networks for RFID-Based Spatial Geometry Inference in Spatial AI Systems]{Graph Neural Networks for RFID-Based Spatial Geometry Inference in Spatial AI Systems}

\author*[1]{\fnm{Curtis L.} \sur{Shull}}\email{cshull@valdosta.edu}

\author[2]{\fnm{Merrick} \sur{Green}}\email{merrick.green@coloradotech.edu}

\author[3]{\fnm{Roy} \sur{Rucker}}\email{roy.rucker@example.com}

\affil*[1]{\orgdiv{College of Computer Science, Engineering, and Technology},
\orgname{Valdosta State University},
\orgaddress{\city{Valdosta}, \state{GA}, \postcode{31698}, \country{USA}}}

\affil[2]{\orgdiv{College of Computer Science, Engineering, and Technology},
\orgname{Colorado Technical University},
\orgaddress{\city{Colorado Springs}, \state{CO}, \postcode{80907}, \country{USA}}}

\affil[3]{\orgdiv{College of Management and Human Potential},
\orgname{Walden University},
\orgaddress{\city{Minneapolis}, \state{MN}, \postcode{55401}, \country{USA}}}

\abstract{%
Indoor spatial understanding remains a fundamental challenge for intelligent systems operating in physical environments. Traditional RFID localization techniques typically estimate positions of tags using signal strength measurements but fail to capture higher-order spatial relationships between objects and infrastructure. Recent work on RFID and wireless indoor localization has increasingly emphasized robust learning under noisy propagation, while recent graph-based localization methods demonstrate the value of relational modeling over isolated samples \cite{liu2023hognnloc,shi2024rfidreview,zhang2024gnnindustry}. This paper introduces a graph-based learning framework that leverages Graph Neural Networks (GNNs) to infer spatial geometry from RFID observations. Rather than predicting isolated coordinates, the proposed system models relationships between RFID readings, antennas, and physical structures within an indoor floorplan. This framing is aligned with recent graph-based indoor positioning and graph construction literature, where topology is a first-class source of information for downstream inference \cite{indoorgnn2023,zhang2024gnnindustry,graphconstruction2023}. The approach integrates signal strength data, floorplan semantics, and spatial constraints into a graph representation where nodes correspond to RFID observations and edges encode proximity and contextual relationships. A GNN is then trained to predict geometric patterns such as linear trajectories, rectangular bounding regions, and movement paths of objects in space. Experiments on RFID data collected in a controlled laboratory environment are used to evaluate the proposed framework and to examine whether graph-based learning can recover meaningful spatial geometries from noisy signal observations. The results illustrate the potential of graph-based spatial reasoning as a foundation for next-generation spatial AI systems, especially when combined with recent floorplan reconstruction and interoperable indoor modeling standards such as IndoorGML 2.0 \cite{yue2023roomformer,ifc2bcm2024,ogcindoorgml2025}. Recent surveys of graph-based wireless localization further support this direction by showing that relational models are increasingly effective when indoor measurements are sparse, noisy, or infrastructure-dependent \cite{wu2023gnnwireless,chen2023graphlocalization}.
}


\begin{document}

\maketitle

\section{Introduction}

Spatial intelligence is essential for robotics, automation systems, and smart environments. Indoor sensing systems must interpret noisy sensor measurements to infer meaningful spatial structure. Recent reviews of RFID indoor localization note that robustness to multipath, occlusion, and environmental variability remains a central limitation of pointwise approaches \cite{shi2024rfidreview,wang2024sensorfusion}.

RFID technology is widely used for tracking assets in indoor environments. Traditional RFID localization methods rely on signal strength measurements to estimate tag positions. However, these approaches treat sensor readings independently and ignore structural information about the environment \cite{shi2024rfidreview}.

Indoor environments contain meaningful spatial relationships including rooms, corridors, equipment locations, and storage containers. Ignoring these relationships limits the ability of sensing systems to reason about spatial geometry. Recent GNN literature has shown that explicitly modeled topology can substantially improve representation quality when relationships among samples are informative \cite{zhang2024gnnindustry,gnnsurvey2023}.

This research reframes RFID localization as a relational spatial inference problem. Instead of predicting isolated coordinates, the system learns patterns in groups of RFID observations and infers geometric structure. This reframing is consistent with broader graph representation learning research, which has shown that structured dependencies among observations can be more informative than independent feature vectors alone \cite{hamilton2023graphrepresentation,li2024spatialgnn}.
Unlike traditional RFID localization methods that estimate independent tag positions, the proposed framework models indoor environments as relational systems composed of interacting entities, including tags, antennas, zones, and containers. The key novelty of this work lies in shifting from point-wise localization to spatial geometry inference, where the objective is to reconstruct structured spatial relationships such as regions, boundaries, and object trajectories. By embedding semantic context and relational dependencies into a graph representation, the model enables higher-order spatial reasoning that is not captured by existing RSSI-based or fingerprinting approaches.

\section{Related Work}

RFID-based spatial analysis has historically been approached from the perspective of localization accuracy. A large portion of the literature focuses on estimating a tag position from received signal strength, phase, time-of-flight, or hybrid signal representations. Recent surveys continue to group the field around fingerprinting, probabilistic methods, and learned regressors, but also identify the need for models that can better exploit spatial context and multi-sensor relationships \cite{shi2024rfidreview,wang2024sensorfusion}. Recent RFID reviews also emphasize that robustness improves when localization pipelines incorporate richer context, cross-sensor evidence, and environment-aware representations rather than relying on signal inversion alone \cite{zhou2023rfidsurvey,kim2024multimodalindoor}.

A second line of work studies indoor sensing as an environmental inference problem. Here, the task is not merely to identify where a tag is at a given instant, but to characterize patterns such as occupancy distributions, object flows, event regions, and spatial usage. RFID is less frequently framed as a modality for recovering structured geometry, although recent wireless localization work using GNNs suggests that graph-based formulations are well suited to noisy indoor measurements \cite{liu2023hognnloc,indoorgnn2023,posgnn2025}.

Graph-based machine learning provides an attractive foundation for this shift in perspective. Graph Neural Networks have become influential because they allow learning over relational structure rather than on isolated samples. A graph can encode which measurements are near each other in space, similar in time, associated with the same antenna, or embedded in the same semantic region. Message passing then aggregates context from neighboring observations. This is useful whenever the signal carried by an individual sample is weak, but the arrangement of many samples contains a more stable pattern \cite{zhang2024gnnindustry,gnnsurvey2023}.

This intuition is also well supported by recent work on graph representation learning and context-aware graph inference, where neighborhood structure improves downstream reasoning under weak or incomplete local evidence \cite{hamilton2023graphrepresentation,park2024contextgnn}.

Within the graph learning literature, most applications have focused on social networks, molecular property prediction, recommendation systems, scene graphs, transport networks, and general relational reasoning. Spatial graphs have also been used in traffic prediction, point cloud analysis, and geospatial learning. These applications share a common principle: topology provides essential information that would be lost in a flat feature matrix. That principle is directly relevant to RFID sensing, where adjacency, density, and local structure often matter more than any single RSSI measurement \cite{zhang2024gnnindustry,graphconstruction2023}.

This paper therefore draws together four strands of research: (i) RFID signal interpretation, (ii) indoor spatial modeling, (iii) graph-based representation learning, and (iv) Spatial AI. The contribution is not to replace established localization methods for all purposes, but to show that when the objective is geometric inference, a graph-centered approach is more aligned with the structure of the problem.

\section{RFID as a Spatial Sensing Infrastructure}

RFID systems are traditionally deployed for identification, inventory control, and presence detection. In many real-world deployments, tags and readers are installed to support operational processes rather than to serve as a dedicated geometric sensing network. However, once observations are aggregated across time and space, the same infrastructure begins to resemble a distributed spatial sensor field. Each detection event provides a partial, uncertain, and noisy hint about where an object might be relative to the antenna layout and surrounding structures \cite{shi2024rfidreview,wang2024sensorfusion}.

This shift in interpretation is important. If one views RFID only as an identification mechanism, then the output of interest is whether a tag has been seen. If one views RFID as a spatial sensing modality, the output of interest becomes the latent spatial arrangement that explains many observed detections. Under this second view, the challenge is not simply signal inversion. Instead, the challenge is to infer the spatial geometry that best accounts for a set of interdependent observations.

Compared with visual sensing, RFID produces much sparser and less intuitive raw data. An image directly reflects visible geometry, while an RFID detection reflects indirect wireless interaction. Nevertheless, RFID has several properties that make it valuable for Spatial AI systems. It can function without line of sight, can operate in visually degraded or occluded environments, and can attach identity to observations in a way that many passive sensing modalities cannot \cite{shi2024rfidreview}.

The limitation is that RFID observations are ambiguous in isolation. A strong RSSI does not map perfectly to a short distance; weak signals can arise from occlusion or orientation effects; repeated detections from the same area can reflect both stationary occupancy and local signal dynamics. For this reason, higher-level spatial understanding must depend on aggregated structure. The system proposed in this paper treats each observation as one element of a relational spatial pattern and uses graph learning to recover the underlying geometry of that pattern.

\section{Spatial AI and Environmental Understanding}

Spatial AI concerns the ability of intelligent systems to perceive, model, and reason about the structure of physical environments. The term is often associated with visual mapping, semantic scene understanding, or embodied perception, but its deeper principle is broader: a system exhibits spatial intelligence when it can infer structured properties of an environment from incomplete evidence and use those properties for downstream reasoning. Recent floorplan reconstruction work has reinforced the value of geometry-aware representations that preserve room polygons, adjacency, and semantics rather than only raster outputs \cite{yue2023roomformer,liu2024cubicasa5k}.

Recent studies on floorplan reasoning and indoor Spatial AI likewise suggest that graph-structured and topology-preserving representations are especially useful when semantic and geometric context must be maintained together \cite{wang2023floorplangnn,liu2025indoorspatialai}.

From this perspective, RFID data offers an unusual but valuable substrate for Spatial AI. The signal observations do not directly depict geometry, yet they are generated by interactions that are fundamentally spatial. Antennas occupy fixed positions. Tags move through rooms, around obstacles, near containers, and across activity zones. Readings cluster in ways that reflect both infrastructure placement and environmental behavior. The environment is therefore encoded implicitly in the data-generating process, even when it is not immediately visible in the raw measurements.

The goal of this work is to convert that implicit spatial information into explicit geometric representations. Rather than predicting a point estimate for each tag independently, the proposed model infers patterns such as elongated trajectories, bounded regions, and dense activity footprints. These patterns are meaningful because they correspond to operational geometry: a path may indicate repeated transport or movement; a rectangular cluster may correspond to container occupancy or region-constrained motion; a compact dense region may reveal persistent usage or storage behavior.

This framing aligns strongly with environmental understanding. The model is not only estimating where a tag might be. It is learning the structure of activity in a spatially instrumented environment. That makes the approach relevant to smart buildings, ambient intelligence, spatial analytics, digital twins, and other domains in which the environment itself is the object of inference.

\section{Problem Formulation}

Let the RFID observation set be denoted by
\[
D = \{o_1,o_2,\dots,o_n\},
\]
where each observation is represented as
\[
o_i = (x_i, y_i, r_i, a_i, t_i, s_i).
\]
Here $(x_i,y_i)$ denotes a spatial coordinate associated with the observation, $r_i$ denotes RSSI or an equivalent signal-strength measurement, $a_i$ denotes the antenna identifier, $t_i$ is the timestamp, and $s_i$ denotes any available semantic context such as zone membership or nearby structural association.

Traditional localization can be written as a mapping
\[
f_{\mathrm{loc}}: o_i \mapsto \hat{p}_i,
\]
where $\hat{p}_i$ is an estimated position. The task addressed in this paper is different. We seek a function
\[
f_{\mathrm{geom}}: D_k \mapsto g_k,
\]
where $D_k \subset D$ is a temporally or contextually grouped subset of observations and $g_k$ is a geometric interpretation of that group. Examples of $g_k$ include a line-like trajectory, a rectangular region, a path, a dense stationary cluster, or a more general shape parameterization.

This formulation emphasizes collective structure. The output is not defined at the level of a single observation, but at the level of a group whose internal relationships contain the spatial information of interest. To support learning on such grouped data, each subset $D_k$ is converted into a graph
\[
G_k=(V_k,E_k,X_k),
\]
where $V_k$ is the set of nodes, $E_k$ is the set of edges, and $X_k$ contains the node features. The geometric inference problem then becomes a graph-to-label or graph-to-parameter prediction task, following recent graph-based indoor localization formulations \cite{liu2023hognnloc,indoorgnn2023}.

\section{Dataset and Floorplan Parsing}

The RFID dataset used in this study consists of structured detections collected in an indoor environment instrumented with antennas and spatially meaningful objects or regions. The exact raw fields depend on the collection pipeline, but the analysis assumes access to at least tag identity, antenna identity, timestamp, and a signal-strength attribute. To move from raw sensor output to spatial inference, the dataset must be contextualized against a floorplan.

The floorplan is not treated as a static background image. It is used as a semantic layer that converts otherwise ambiguous coordinates into interpretable environmental relationships. A point can be described not only by its $(x,y)$ value but also by whether it falls inside a room, near a container, within a traffic corridor, or in the influence region of a given antenna. This transformation is crucial because spatial inference depends on more than geometric distance. Two observations that are near each other numerically may have very different meanings if they lie on opposite sides of a wall or in functionally different parts of the environment. Recent floorplan reconstruction systems and indoor information models similarly emphasize explicit room boundaries, polygonal structure, and topology-aware semantics \cite{yue2023roomformer,liu2024cubicasa5k,ogcindoorgml2025}.

This emphasis on explicit structural relationships is also reflected in recent graph-based floorplan modeling work, where topological consistency improves downstream spatial reasoning and representation quality \cite{wang2023floorplangnn}.

The parsing process therefore extracts several classes of information from the floorplan:
\begin{enumerate}[label=\arabic*.]
\item structural boundaries such as rooms, corridors, or enclosing regions,
\item infrastructure positions such as antenna locations,
\item object-linked zones such as containers, shelves, or storage footprints,
\item semantic labels that can be attached to observations during feature generation.
\end{enumerate}

Once this spatial layer has been built, the RFID observations can be enriched in a way that supports relational learning.
A detection near a container is no longer just a point with RSSI;
it becomes a point with context.
This addition makes the eventual graph representation far more expressive.

The implementation uses Python with PyTorch, PyTorch Geometric, pandas, NumPy, scikit-learn, matplotlib, Pillow, and XML parsing utilities.
The indoor floorplan is parsed from a \texttt{.flp} file, from which zones, antenna polygons, crate polygons, centers, and optional background imagery are extracted.
The RFID table is filtered to records with valid spatial coordinates, signal strength, floorplan identity, and container identity.
Only observations associated with the parsed floorplan are retained for modeling.
This floorplan-aware filtering is important because it ensures that semantic and geometric enrichment are defined in a consistent coordinate frame.

\section{Dataset Construction and Sample Generation}

A major methodological step in this work is converting a continuous stream of RFID detections into graph samples that can be used for supervised or semi-supervised learning.
This sample generation process determines the granularity at which spatial patterns are learned.

Observations are first grouped into time windows or event-defined segments.
Let
\[
T_k = \{o_i \mid t_i \in [k\Delta,(k+1)\Delta)\}
\]
denote the observations collected during a time interval of width $\Delta$.
Alternative grouping criteria can incorporate object identity, antenna transitions, or operational sessions.
The central idea is that each group should correspond to a local spatial episode whose geometry can be meaningfully inferred.

After temporal grouping, filtering is applied to remove groups with insufficient signal support.
Very small groups do not contain enough structure for robust graph construction and are therefore either merged, discarded, or handled separately.
For the remaining groups, candidate geometric labels can be generated in several ways:
rule-based fitting, manual review, clustering-derived shape descriptors, or weak supervision from environmental priors.
In practice, this study assumes graph samples associated with interpretable geometry categories such as trajectory-like, rectangular-region, or irregular-path.

This sample construction stage has both methodological and scientific importance.
Methodologically, it creates the graph units consumed by the learning model.
Scientifically, it defines what kind of geometry the system is expected to recover.
A poorly chosen grouping strategy can obscure meaningful patterns or blend distinct spatial episodes.
A carefully chosen strategy, by contrast, reveals structure that would be invisible in pointwise analysis.

In the implemented pipeline, observations are grouped using a fixed temporal window of
\[
\Delta = 5.0 \text{ seconds}.
\]
For each record, the fields \texttt{TagFirstDiscovered} and \texttt{TagLastSeen} are converted from mm:ss format into seconds, and a discrete time bucket is assigned as
\[
b_i = \left\lfloor \frac{t_i}{\Delta} \right\rfloor.
\]
Grouped samples are then formed using the tuple
\[
(\text{FloorplanUId}, \text{ContainerId}, \text{time\_bucket}, \text{TagUId}),
\]
followed by a second grouping stage over
\[
(\text{FloorplanUId}, \text{ContainerId}, \text{time\_bucket})
\]
to create graph-level samples.

Within each grouped sample, repeated detections are aggregated by taking the mean of spatial and signal attributes such as $X$, $Y$, and RSSI, while count-related variables such as \texttt{TagCount} are summed.
Duration-based attributes are computed from the difference between first-seen and last-seen timestamps.
Groups with fewer than two aggregated observations are discarded because they do not support meaningful graph construction or geometric inference.

To generate weak supervision for the geometry labels, a simple rule-based scheme is used.
Let $P_k$ denote the set of points in sample $k$ and let $n_k = |P_k|$.
If $n_k=2$, the sample is labeled as a line-like geometry.
If $n_k \ge 4$ and the spatial extent satisfies
\[
(x_{\max}-x_{\min}) > 10
\quad \text{and} \quad
(y_{\max}-y_{\min}) > 10,
\]
the sample is labeled as rectangular.
All other valid samples are labeled as path-like.
Although simple, this procedure provides a practical weakly supervised target generation strategy for early-stage experimentation.

\section{Feature Engineering}

Feature engineering is essential because the learning model must integrate heterogeneous signals: coordinates, sensor readings, structural context, and relational cues.
Each observation is transformed into a feature vector
\[
x_i = [x_i, y_i, r_i, d^{(a)}_i, d^{(c)}_i, d^{(z)}_i, \phi_i],
\]
where $d^{(a)}_i$ is the distance to the nearest antenna or to the relevant antenna, $d^{(c)}_i$ is the distance to the nearest container or object region, $d^{(z)}_i$ is a zone-based distance or region indicator, and $\phi_i$ may include additional engineered variables such as normalized timestamp, antenna one-hot encoding, local density, or moving-window statistics.

In the implemented notebook, each node feature vector contains the following attributes:
\begin{align}
x_i = [&X_i,\; Y_i,\; \mathrm{RSSI}_i,\; \mathrm{TagCount}_i,\; \mathrm{duration}_i, \\
      &\mathrm{AntennaEnc}_i,\; \mathrm{ZoneEnc}_i,\; \mathrm{ContainerEnc}_i, \\
      &\mathrm{InsideZone}_i,\; \mathrm{DistZoneCenter}_i, \\
      &\mathrm{DistNearestAntenna}_i,\; \mathrm{InsideCrate}_i,\; \mathrm{DistNearestCrate}_i]
\end{align}

Here, \(\mathrm{AntennaEnc}_i\), \(\mathrm{ZoneEnc}_i\), and \(\mathrm{ContainerEnc}_i\) are integer encodings of categorical identifiers obtained using label encoding.
The binary variable \(\mathrm{InsideZone}_i\) indicates whether the observation falls inside the polygon of its associated zone.
The quantity \(\mathrm{DistZoneCenter}_i\) measures Euclidean distance to the center of the assigned zone.
Similarly, \(\mathrm{DistNearestAntenna}_i\) and \(\mathrm{DistNearestCrate}_i\) measure distances from the observation to the nearest antenna center and the nearest crate or container center, respectively.
The variable \(\mathrm{InsideCrate}_i\) indicates whether the point lies inside any parsed crate polygon.

These features were chosen to combine three forms of information:
measurement strength,
relative location with respect to sensing infrastructure,
and semantic environmental context extracted from the floorplan.
This combination makes the input representation substantially richer than raw coordinates and RSSI alone.

The spatial coordinates serve as the base geometric descriptor.
However, coordinates alone do not explain how an observation relates to the sensing infrastructure.
RSSI helps to encode signal intensity, but RSSI is noisy and nonlinear.
Distance-to-antenna features help stabilize this information by grounding detections relative to known infrastructure.
Similarly, distance-to-container and zone features introduce environmental semantics that are useful for distinguishing free movement from structured occupancy.

Additional derived features can substantially improve representation quality.
Examples include:
\begin{itemize}
\item local neighbor count within a radius,
\item variance of RSSI in a short temporal neighborhood,
\item angle or direction from the nearest antenna,
\item binary flags indicating whether a point lies inside a corridor, room, or storage footprint,
\item normalized path order within a temporal sequence.
\end{itemize}

These features are not arbitrary embellishments.
They operationalize the hypothesis that geometry emerges from context.
A point is easier to interpret when one knows whether it is central to a region, on a boundary, near an infrastructure element, or part of a temporally coherent sequence.
The resulting feature matrix therefore serves as a compact representation of both measurement and environment.

\section{Graph Construction}

For each grouped sample, a graph
\[
G=(V,E)
\]
is constructed. Each node corresponds to an RFID observation. The central design question is how to define the edges. Because the objective is spatial geometry inference, edges should connect observations that are likely to participate in a shared geometric pattern.

The simplest construction uses distance-threshold adjacency:
\[
A_{ij} =
\begin{cases}
1 & \|p_i-p_j\| < \tau,\\
0 & \text{otherwise}.
\end{cases}
\]
This approach produces a graph in which local spatial neighborhoods become connected substructures.
It is computationally straightforward and often effective, especially when the observation density is moderate and the environment is not extremely fragmented.

In practice, richer edge definitions may be preferable. For example, one may connect nodes if they are close in both space and time, or if they share the same antenna context. Edges can also be weighted according to distance, signal similarity, or semantic compatibility. A weighted adjacency matrix can be written as
\[
w_{ij} = \exp\left(-\frac{\|p_i-p_j\|^2}{2\sigma^2}\right)\cdot \eta_{ij},
\]
where $\eta_{ij}$ may encode time-window similarity, zone consistency, or antenna agreement. Such weighting gives the model a softer notion of neighborhood and can improve stability when the point distribution is uneven.

The importance of graph construction cannot be overstated. The GNN does not discover relationships from scratch; it learns over the topology provided. If the graph poorly reflects the latent spatial organization, then message passing will propagate irrelevant information. If the graph aligns with the geometry of the data, then the model can aggregate exactly the type of context needed for inference \cite{graphconstruction2023}.

Recent work on graph construction for spatial learning reinforces this point by showing that edge design is not merely a preprocessing detail but a major determinant of downstream model quality and interpretability \cite{xu2023graphconstruction}.

In the implemented graph construction procedure, an edge is added between two nodes \(i\) and \(j\) if their Euclidean spatial distance is below a fixed threshold:
\[
\|p_i - p_j\| < \tau,
\qquad \tau = 120.0.
\]
The threshold is specified in the same coordinate system as the floorplan.

For every directed edge \((i,j)\), the edge attribute vector is defined as
\[
e_{ij} =
[
d_{ij},\,
|\mathrm{RSSI}_i - \mathrm{RSSI}_j|,\,
\mathbf{1}(a_i=a_j),\,
\mathbf{1}(z_i=z_j),\,
\mathbf{1}(c_i=c_j)
],
\]
where \(d_{ij}\) is Euclidean distance, \(a_i\) is antenna identity, \(z_i\) is zone identity, and \(c_i\) is container identity.
Thus, the graph encodes not only proximity but also whether neighboring observations share common sensing or semantic context.

If a grouped sample produces no valid edges under the threshold rule, it is excluded from the training dataset.
This design decision avoids degenerate graphs and ensures that each training example contains at least minimal relational structure.

\section{Mathematical Formulation of Spatial Graph Learning}

Given a graph sample $G=(V,E,X)$ with $|V|=N$ nodes and node feature matrix
\[
X \in \mathbb{R}^{N\times d},
\]
the objective is to learn a function from graph structure to geometric interpretation.
Let $A$ denote the adjacency matrix and $\tilde{A}=A+I$ the adjacency matrix with self-loops.
Let $\tilde{D}$ be the diagonal degree matrix of $\tilde{A}$.
A normalized propagation operator can then be defined as
\[
\hat{A} = \tilde{D}^{-1/2}\tilde{A}\tilde{D}^{-1/2}.
\]

A standard graph convolution layer takes the form
\[
H^{(\ell+1)} = \sigma\left(\hat{A}H^{(\ell)}W^{(\ell)}\right),
\]
with $H^{(0)}=X$, learnable weights $W^{(\ell)}$, and nonlinearity $\sigma$.
This operation can also be described node-wise:
\[
h_i^{(\ell+1)} = \sigma\left(
\sum_{j\in \mathcal{N}(i)\cup\{i\}}
\frac{1}{\sqrt{\tilde{d}_i\tilde{d}_j}}
W^{(\ell)}h_j^{(\ell)}
\right),
\]
where $\mathcal{N}(i)$ is the neighborhood of node $i$ and $\tilde{d}_i$ is the degree of node $i$ in the self-loop-augmented graph.

This formulation matters because it formalizes contextual aggregation.
Each updated node embedding is a transformed average of its own representation and those of its neighbors.
If neighboring observations belong to the same geometric structure, then message passing amplifies the shared pattern.
If they do not, then the graph design or learned weights should suppress irrelevant influence.

After $L$ layers, the node embeddings encode increasingly broader relational context.
A graph-level embedding can be obtained using pooling:
\[
z = \mathrm{POOL}\left(\{h_i^{(L)}\}_{i=1}^{N}\right).
\]
Mean pooling is often effective:
\[
z = \frac{1}{N}\sum_{i=1}^{N} h_i^{(L)}.
\]
Max pooling or attention-based pooling may also be used when only a subset of nodes is geometrically informative.

The graph embedding is then passed to downstream heads.
For geometry classification:
\[
\hat{y} = \mathrm{softmax}(W_c z + b_c).
\]
For geometry parameter regression:
\[
\hat{g} = W_r z + b_r.
\]
A multitask objective combines both:
\[
\mathcal{L} = \mathcal{L}_{\mathrm{class}} + \lambda \mathcal{L}_{\mathrm{geom}},
\]
where $\lambda$ balances classification and regression.

This formalization aligns tightly with the scientific goal.
Rather than asking the model to infer geometry from isolated vectors, it requires the model to reason through the relational structure of observations.
That is the central conceptual shift of the paper. Related context-aware GNN research supports this formulation by showing that explicitly relational inference mechanisms are particularly valuable when geometry must be recovered from partial local observations rather than directly measured state variables \cite{park2024contextgnn}.

\section{Graph Neural Network Architecture}

The architecture used in this work is intentionally interpretable. It consists of a stack of graph convolution layers followed by graph pooling and separate output heads. A representative instance uses three graph convolution layers with hidden dimension $64$:
\begin{align}
H^{(1)} &= \sigma(\hat{A}XW^{(1)}),\\
H^{(2)} &= \sigma(\hat{A}H^{(1)}W^{(2)}),\\
H^{(3)} &= \sigma(\hat{A}H^{(2)}W^{(3)}).
\end{align}

The implemented model uses the Graph Convolutional Network (GCN) operator from PyTorch Geometric.
Specifically, three graph convolution layers are applied:
\begin{align}
H^{(1)} &= \mathrm{ReLU}(\mathrm{GCNConv}(X, E)),\\
H^{(2)} &= \mathrm{ReLU}(\mathrm{GCNConv}(H^{(1)}, E)),\\
H^{(3)} &= \mathrm{ReLU}(\mathrm{GCNConv}(H^{(2)}, E)).
\end{align}
Each hidden layer has dimension \(64\).
After message passing, global mean pooling produces a graph-level embedding.

Two multilayer perceptron heads are then used.
The classification head predicts one of three geometry classes:
\[
\{\text{line},\text{rect},\text{path}\}.
\]
The regression head predicts an 8-dimensional geometry parameter vector.
For line samples, the first four entries represent two endpoints.
For rectangular samples, the first four entries represent \((x,y,w,h)\).
For path-like samples, up to four ordered \((x,y)\) points are flattened into an 8-dimensional vector, with zero-padding when fewer than four ordered points are available.

The first layer mainly integrates local signal and semantic context.
The second layer begins to encode neighborhood structure and short-range pattern consistency.
The third layer captures broader subgraph-level shape information.
This progression is conceptually useful because many target geometries, such as trajectories and bounded regions, are not visible at the level of one or two points but emerge over several relational hops.

A global mean pooling layer transforms the final node embeddings into a graph representation:
\[
z = \frac{1}{|V|}\sum_{i=1}^{|V|} h_i^{(3)}.
\]
The pooled vector summarizes the sample and is passed to:
\begin{itemize}
\item a classification head for geometry type,
\item a regression head for geometry parameters such as length, bounding-box extent, orientation, or path descriptors.
\end{itemize}

Dropout, layer normalization, and residual links may be added depending on the scale and variability of the dataset.
If class imbalance is significant, focal loss or weighted cross-entropy can be used in the classification head.
If the target geometry is highly structured, the regression head can be replaced by a parameter decoder tailored to specific shapes.

The architecture is intentionally general enough to support multiple geometry tasks while remaining closely tied to the spatial meaning of the data.
It does not assume dense sensor coverage or perfect signal calibration.
Instead, it exploits the topology and semantics embedded in the observation graph.

\section{Training Objective and Optimization}

The implemented training procedure uses a joint loss consisting of cross-entropy for geometry classification and mean squared error for geometry regression.
Let \(\hat{y}\) denote the predicted class logits and \(\hat{g}\) the predicted geometry parameter vector.
The loss is
\[
\mathcal{L}
=
\mathcal{L}_{\mathrm{CE}}(\hat{y}, y)
+
\mathcal{L}_{\mathrm{MSE}}(\hat{g}, g).
\]
In the current implementation, the weighting coefficient is fixed to \(1\), so both terms contribute additively without additional scaling.

Optimization is performed with Adam using learning rate
\[
\eta = 0.001.
\]
Training is run for \(20\) epochs with batch size \(32\).
The dataset is split at graph level using a \(20\%\) held-out test partition, after which \(10\%\) of the remaining training set is used for validation.
In other words, the split is performed in two stages:
first train/test, then train/validation within the training portion.
This prevents direct sample leakage across graph instances while preserving enough data for validation and testing.

The model is trained on GPU when available and otherwise on CPU.
Training and validation losses are tracked across epochs, and the notebook generates a learning-curve plot for diagnostic inspection.

An important property of this training setup is that it encourages the model to learn both \emph{what kind} of geometry is present and \emph{how} that geometry is instantiated.
This is stronger than pure classification because it pushes the latent graph embedding to capture meaningful structure instead of learning a coarse decision boundary only.

\section{Algorithmic Pipeline}

\begin{algorithm}[!t]
\caption{RFID-Based Spatial Geometry Inference}
\begin{algorithmic}[1]
\STATE Load RFID detections and floorplan metadata
\STATE Parse antenna locations, zones, and structural objects
\STATE Group observations into temporal or event-based samples
\FOR{each sample}
    \STATE Compute enriched node features
    \STATE Construct a spatial or spatio-temporal graph
    \STATE Assign or retrieve target geometry labels and parameters
\ENDFOR
\STATE Split graph samples into train, validation, and test sets
\STATE Train graph neural network with multitask loss
\STATE Evaluate classification and geometric parameter accuracy
\STATE Generate visualization outputs for interpretation and error analysis
\end{algorithmic}
\end{algorithm}

This algorithmic view clarifies the relationship between data engineering and learning.
The graph model is only one part of the overall system.
Equally important are the floorplan parser, feature generator, graph constructor, and visualization layer.
Together these stages form a complete spatial inference pipeline rather than a single predictive model.

\section{Visualization Pipeline}

Visualization is central to this work because the scientific claim is about \emph{spatial structure}.
A successful method should not only produce quantitative improvements, but also generate outputs that can be interpreted in geometric terms.
The visualizations therefore serve three purposes:
\begin{enumerate}[label=\arabic*.]
\item to describe the raw data and its spatial characteristics,
\item to expose how graph construction organizes the observations,
\item to reveal whether model predictions align with human-interpretable geometry.
\end{enumerate}

This section expands each visualization beyond simple captioning.
For every figure, the key question is not just what is displayed, but why that display matters to the argument of the paper.

\subsection{Floorplan and Infrastructure View}

\begin{figure}[!t]
\centering
\includegraphics[width=0.84\linewidth]{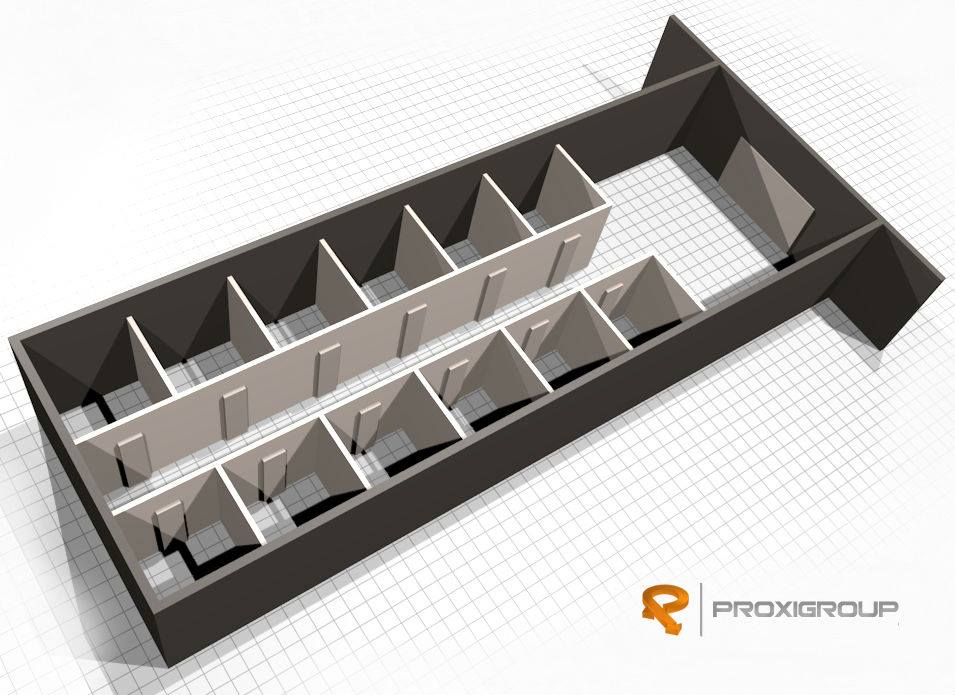}
\caption{Floorplan-level view of the environment, including spatial infrastructure that provides geometric context for RFID observations.}
\label{fig:floorplan}
\end{figure}

The floorplan visualization is the conceptual anchor of the entire study.
It shows the physical environment in which the RFID detections occur and makes clear that the sensing problem is embedded in a structured space rather than in an abstract coordinate plane.
A raw table of RFID detections does not express which points lie near walls, containers, or specific infrastructure.
The floorplan does.

Its relevance is therefore foundational.
First, it motivates the use of semantic features.
When one can see the antenna layout and environmental structure, it becomes obvious that the meaning of a detection depends partly on where it occurs relative to those fixed elements.
Second, the floorplan provides a reference frame for all subsequent visualizations.
Heatmaps, graphs, trajectories, and predictions become interpretable only because they can be read against a shared spatial background.
Third, it demonstrates that the proposed problem is not arbitrary coordinate prediction but environmental understanding.
The objective is to infer geometry in a meaningful physical layout.

In practical terms, this figure helps explain why two observations with similar RSSI can have very different implications.
If one is located in a corridor-like region and another beside a container cluster, their spatial roles are different.
The model must learn such distinctions, and the floorplan shows the source of that structure.

\subsection{RFID Signal Heatmap}

\begin{figure}[!t]
\centering
\includegraphics[width=0.84\linewidth]{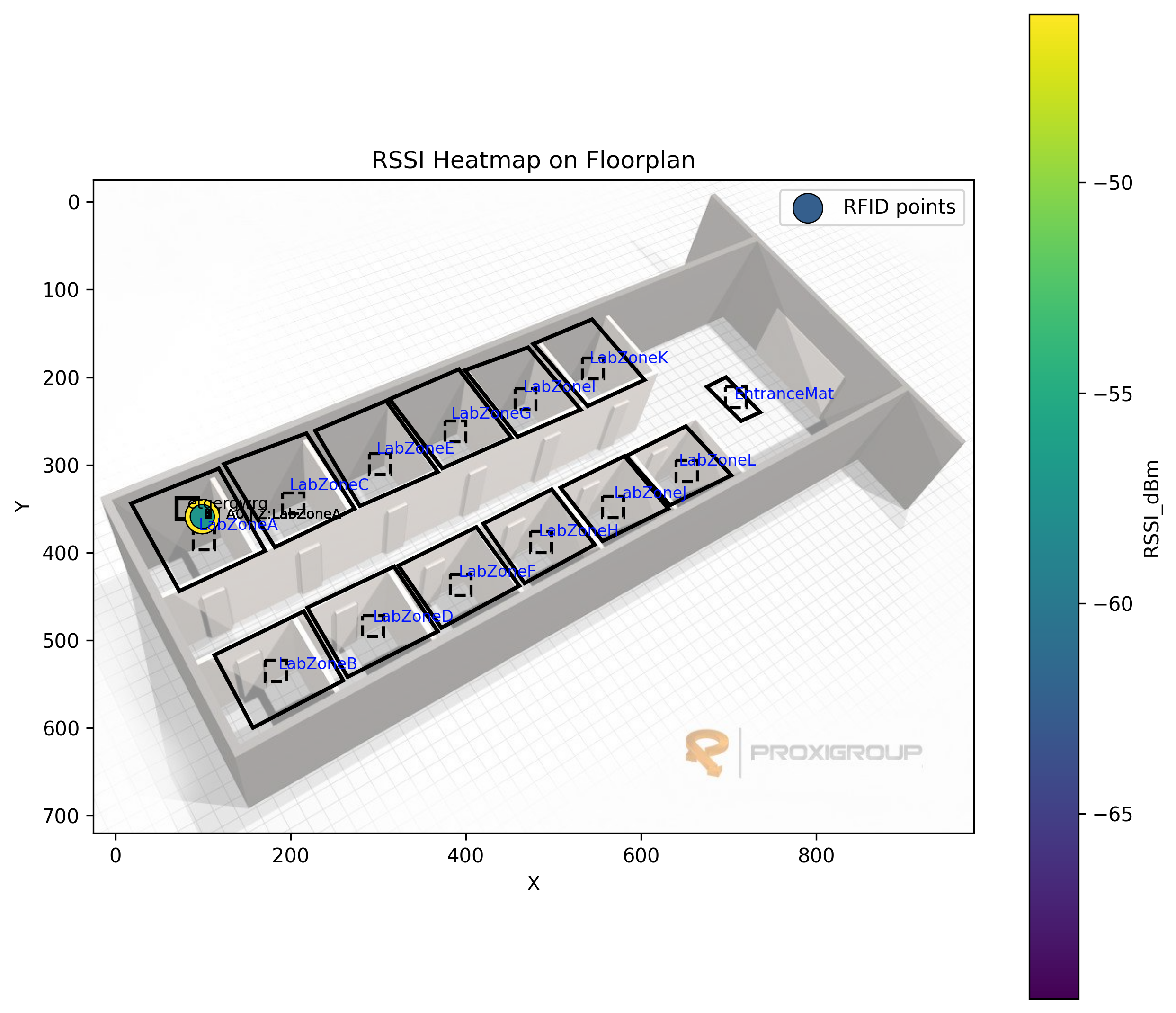}
\caption{RSSI heatmap showing the spatial distribution of signal intensity across the environment.}
\label{fig:rssiheatmap}
\end{figure}

The RSSI heatmap visualizes how signal intensity varies across the environment.
Its relevance lies in revealing the complexity of the raw sensing field.
In an idealized localization setting, signal strength would map smoothly to distance and therefore to location.
The heatmap shows why such an assumption is rarely adequate in real indoor environments.

Patterns in the heatmap often reflect several overlapping factors:
distance from antennas, antenna orientation, reflection and absorption effects, structural occlusion, and repeated operational activity in specific locations.
As a result, the same approximate signal intensity can appear in multiple places, and neighboring positions can exhibit different RSSI values.
This is exactly why a relational model is needed.
If signal is spatially irregular, then the inference process must rely on context and pattern, not on single-point inversion.

The heatmap also provides evidence for the idea that RFID sensing captures more than raw distance.
Localized ``hot'' regions may indicate persistent tag activity, stable occupancy, or repeated movement through particular areas.
Diffuse regions may reflect transitional paths or signal uncertainty.
When read alongside the floorplan, the heatmap can reveal whether strong signal zones align with infrastructure and whether weak signal areas correspond to shadowed or peripheral regions.

From an interpretability standpoint, this figure justifies the entire modeling decision.
It visually argues that the data generating process is too complex for pointwise heuristics alone, and that geometry must be inferred from structured patterns in the field.

\subsection{Observation Density Heatmap}

\begin{figure}[!t]
\centering
\includegraphics[width=0.84\linewidth]{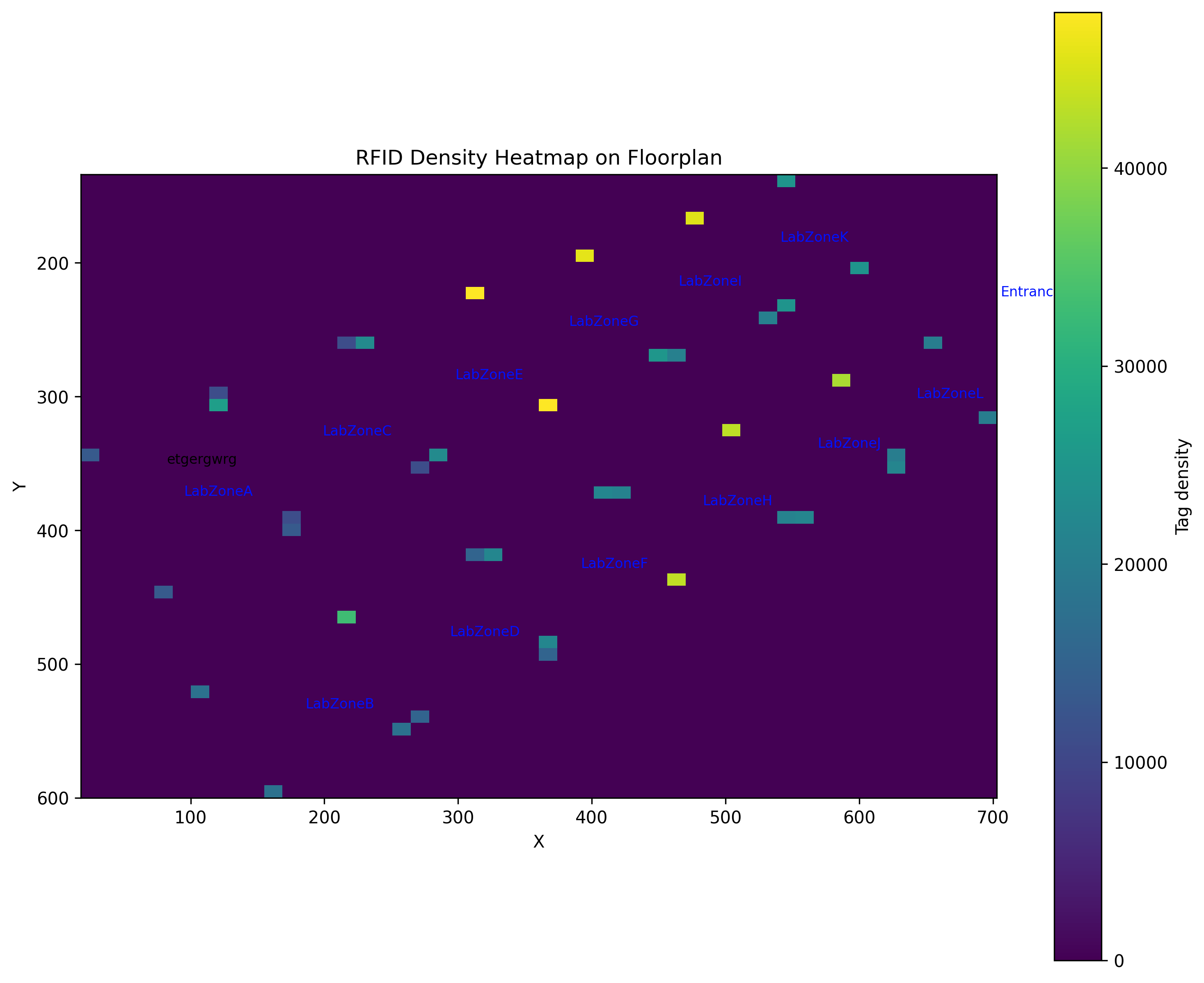}
\caption{Density heatmap of RFID observations, highlighting persistent activity regions and spatial concentration of detections.}
\label{fig:densityheatmap}
\end{figure}

Where the RSSI heatmap emphasizes signal magnitude, the density heatmap emphasizes \emph{where observations accumulate}.
This distinction is important.
A region with high observation density may or may not have the highest signal values, but it often corresponds to repeated operational relevance:
stored items, recurrent transit paths, or zones where detections naturally cluster because of the environment and workflow.

The density visualization is relevant for three reasons.
First, it reveals the empirical support available to the model in different parts of the environment.
Areas with high density provide rich relational structure, while low-density areas are harder to model robustly.
Second, it helps identify whether target geometries are likely to be stable or under-sampled.
If a trajectory repeatedly traverses a region, its geometric signature is easier to learn than if it appears only once in a sparse corner.
Third, the density map can expose data imbalance in spatial terms, complementing standard class imbalance analysis.

This figure also supports the interpretation of later results.
If the model performs best in high-density regions and less reliably in sparse zones, the density heatmap provides a principled explanation.
It transforms a generic performance observation into a spatially grounded insight.

\subsection{Graph Topology Visualization}

\begin{figure}[!t]
\centering
\includegraphics[width=0.84\linewidth]{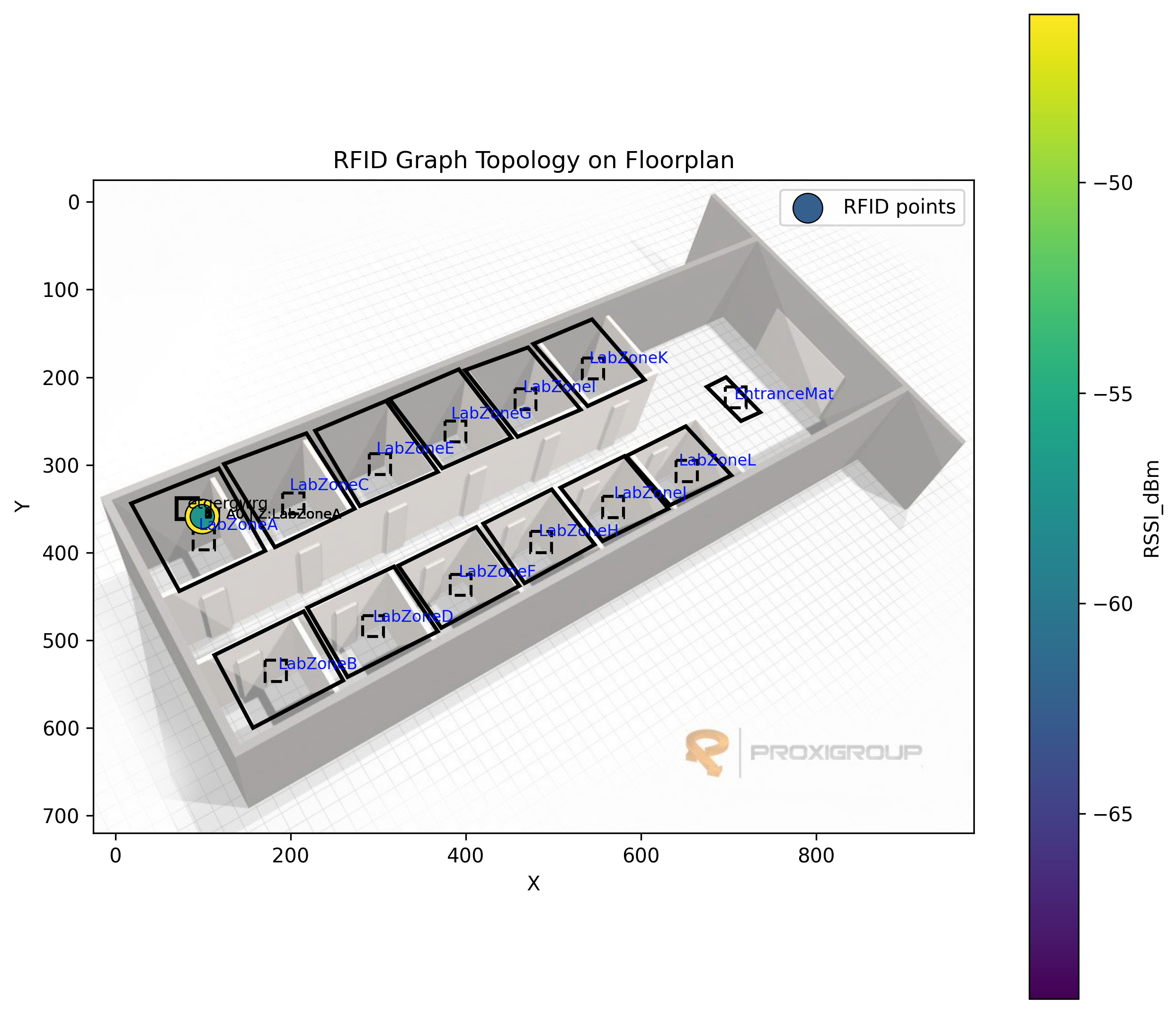}
\caption{Graph topology induced from RFID observations, where nodes are detections and edges encode spatial or contextual proximity.}
\label{fig:graphtopology}
\end{figure}

The graph topology visualization is the bridge between raw data and the learning model.
It makes the abstract graph representation concrete by showing how individual observations become part of a relational structure.
Its importance is methodological and conceptual.

Methodologically, this figure verifies that the graph construction process produces plausible neighborhoods.
If the graph is too fragmented, message passing will fail to integrate enough context.
If it is too dense, local geometric structure will be blurred by irrelevant edges.
The topology visualization therefore acts as a diagnostic tool for the graph-building stage.

Conceptually, the figure demonstrates how geometry can be represented relationally.
A trajectory-like pattern does not need to be encoded as a single line object in advance.
It can emerge from a chain or cluster of connected observations.
A region-like pattern does not need to be hard-coded as a rectangle.
It can appear as a dense subgraph occupying a bounded area.
This is the key reason graph learning is appropriate for the problem: topology allows geometry to emerge from local relationships.

When viewed qualitatively, connected components, elongated subgraphs, dense cores, and branching structures can all correspond to meaningful spatial phenomena.
The figure therefore provides an interpretable explanation of what the GNN is actually operating on.

\subsection{Trajectory Visualization}

\begin{figure}[!t]
\centering
\includegraphics[width=0.84\linewidth]{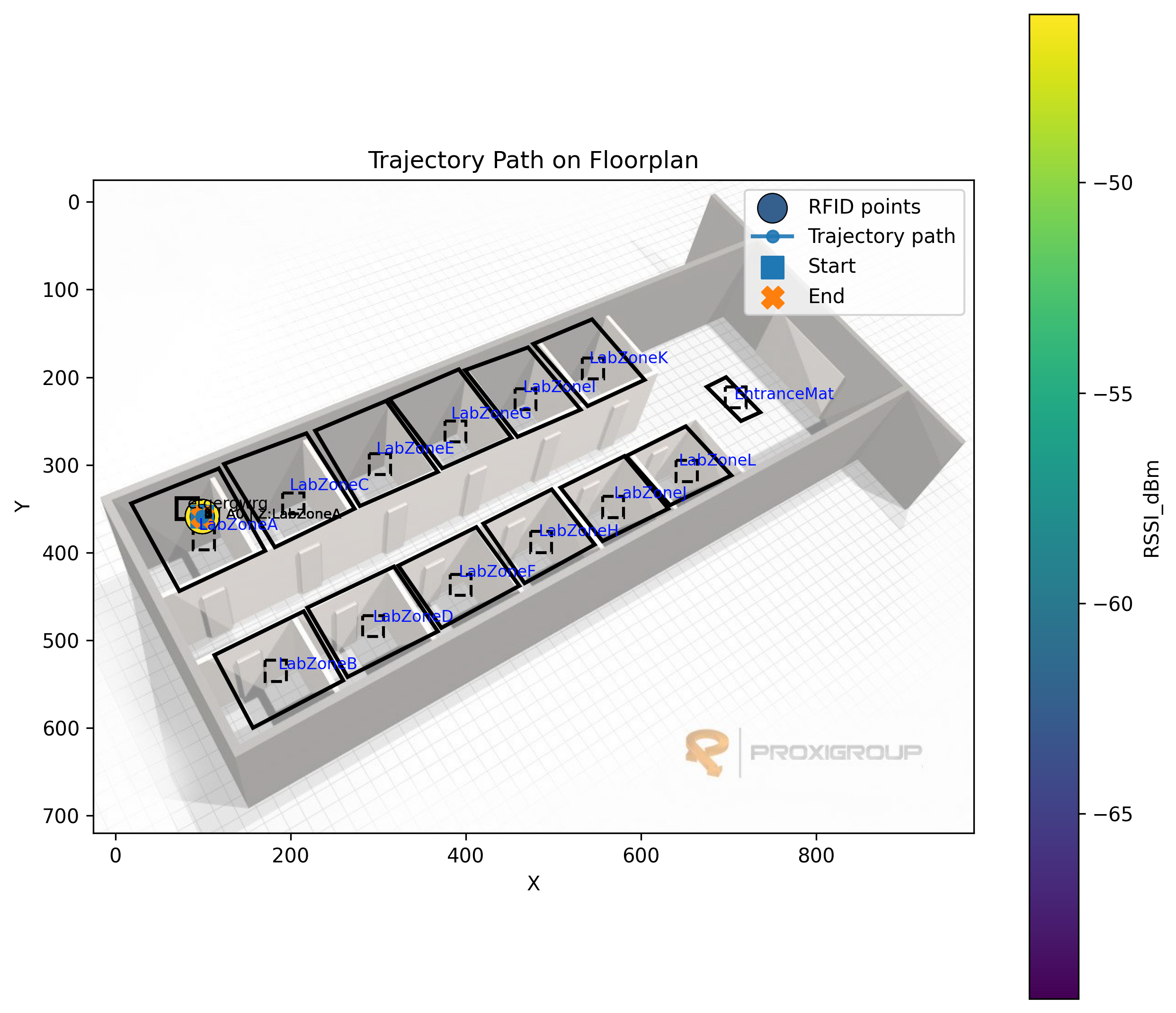}
\caption{Example trajectory inferred from a temporally ordered sequence of RFID observations.}
\label{fig:trajectory}
\end{figure}

The trajectory visualization shows how a sequence of detections can be interpreted as motion through space.
Its relevance is direct: trajectories are among the clearest examples of spatial geometry that cannot be reduced to independent coordinate predictions.
A trajectory has continuity, directionality, and path structure.
Those properties exist only at the level of multiple observations.

This figure helps explain why temporal grouping is scientifically meaningful.
If observations are grouped appropriately, the model can learn patterns of sequential spatial continuity.
Even when timestamps are not explicitly used in the GNN message function, the graph can still preserve motion structure through sample construction and feature design.
The resulting path-like graph is then recognizable both visually and computationally.

The trajectory view is also important for application relevance.
Many environments care less about a precise instantaneous point and more about where an object moved, whether it followed an expected path, or whether it entered a region it normally would not.
By showing that RFID-based geometry inference can recover movement paths, the figure broadens the practical scope of the method beyond static localization.

\subsection{Predicted Geometry Overlay}

\begin{figure}[!t]
\centering
\includegraphics[width=0.84\linewidth]{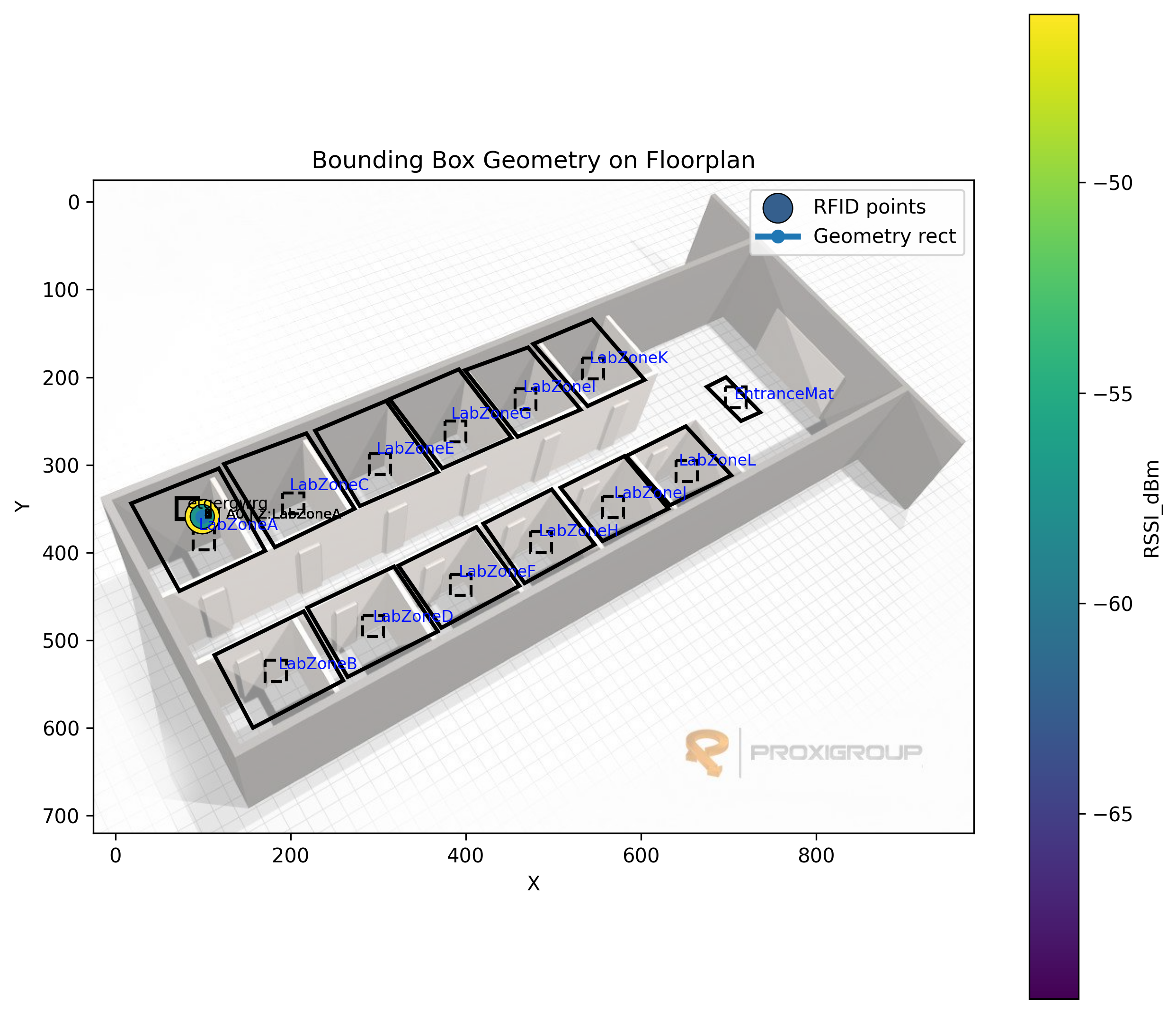}
\caption{Predicted geometry overlaid on observed RFID data, enabling direct comparison between inferred structure and measurement support.}
\label{fig:predictionoverlay}
\end{figure}

The predicted geometry overlay is one of the most important interpretability figures in the paper.
It places the model output directly on top of the observations that generated it.
This allows the reader to assess whether the prediction is not just numerically plausible but spatially sensible.

Its relevance is threefold.
First, it provides qualitative validation.
If the predicted line, path, or region aligns with the support implied by the observation cluster, then the model is capturing real structure.
Second, it reveals failure modes.
Overly large predicted regions, fragmented paths, or misaligned orientations become visually obvious, which is especially useful when aggregate metrics hide localized errors.
Third, the overlay figure connects quantitative results to operational meaning.
A high classification accuracy is useful, but a geometry overlay shows what that success looks like in physical space.

This figure is also central to communicating the contribution of the paper.
The novelty of the method lies in inferring geometry from RFID observations.
The overlay demonstrates that claim in the most direct possible way.

\subsection{Embedding Heatmap}

\begin{figure}[!t]
\centering
\includegraphics[width=0.84\linewidth]{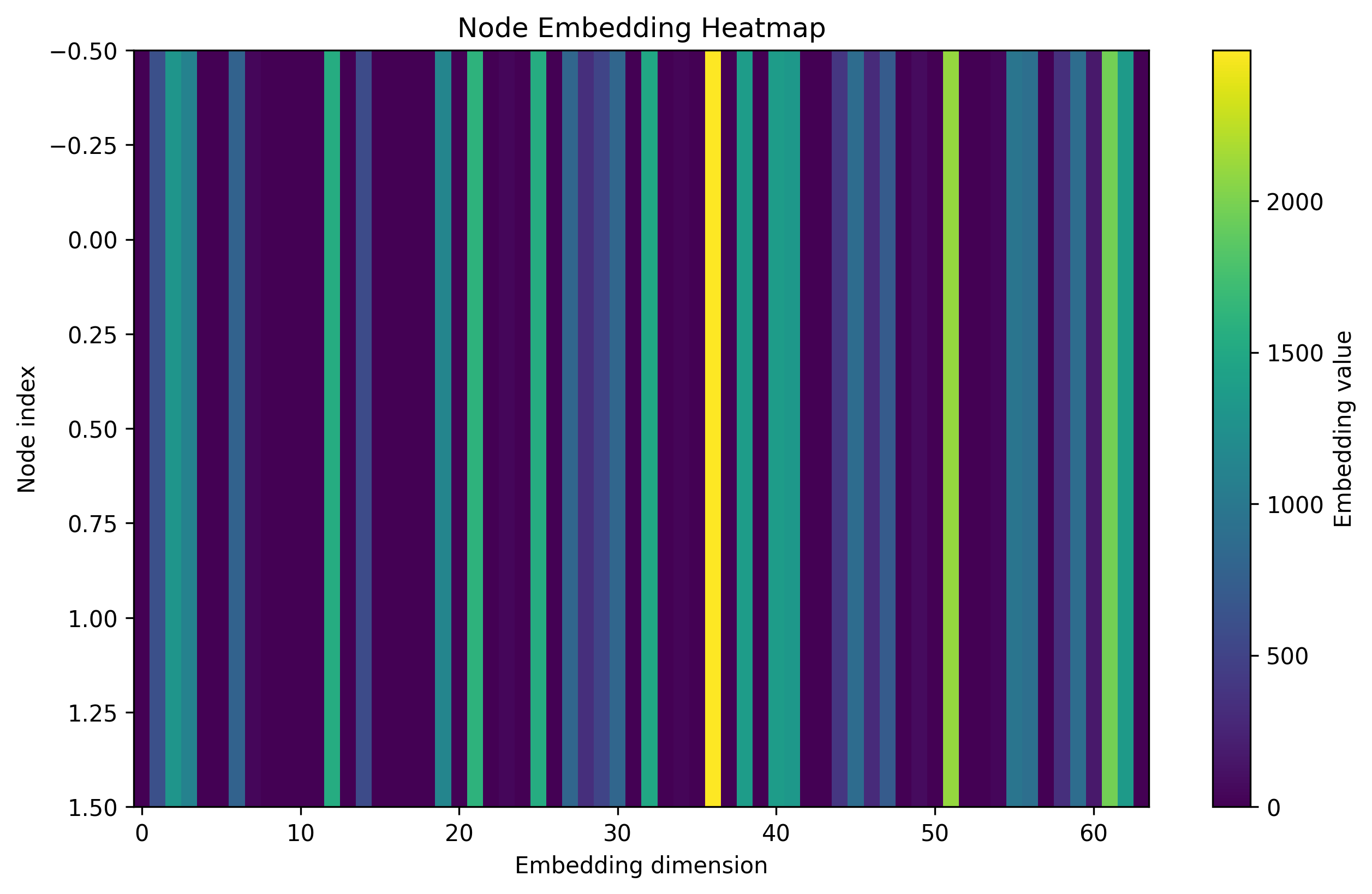}
\caption{Heatmap of learned node or graph embeddings, illustrating internal representation patterns induced by the GNN.}
\label{fig:embeddingheatmap}
\end{figure}

The embedding heatmap provides a window into the internal representation learned by the GNN.
Unlike the previous visualizations, which remain in physical space, this figure operates in latent feature space.
Its relevance lies in interpretability at the representation level.

If the model is learning useful relational structure, then embeddings associated with similar geometric contexts should develop recognizable regularities.
Nodes belonging to the same trajectory or region may occupy similar latent patterns.
Graphs corresponding to similar geometry classes may cluster in embedding space.
The heatmap makes this behavior visible by showing how activation patterns vary across nodes or samples.

This figure is valuable because it demonstrates that the network is not acting as a black-box coordinate regressor.
It is constructing internal representations that organize the data according to geometry-relevant features.
That supports the broader claim that graph-based spatial reasoning is occurring inside the model, not merely at the level of final outputs.

\subsection{RFID Observations on the Proxilab Floorplan}

\begin{figure}[!t]
\centering
\includegraphics[width=0.85\linewidth]{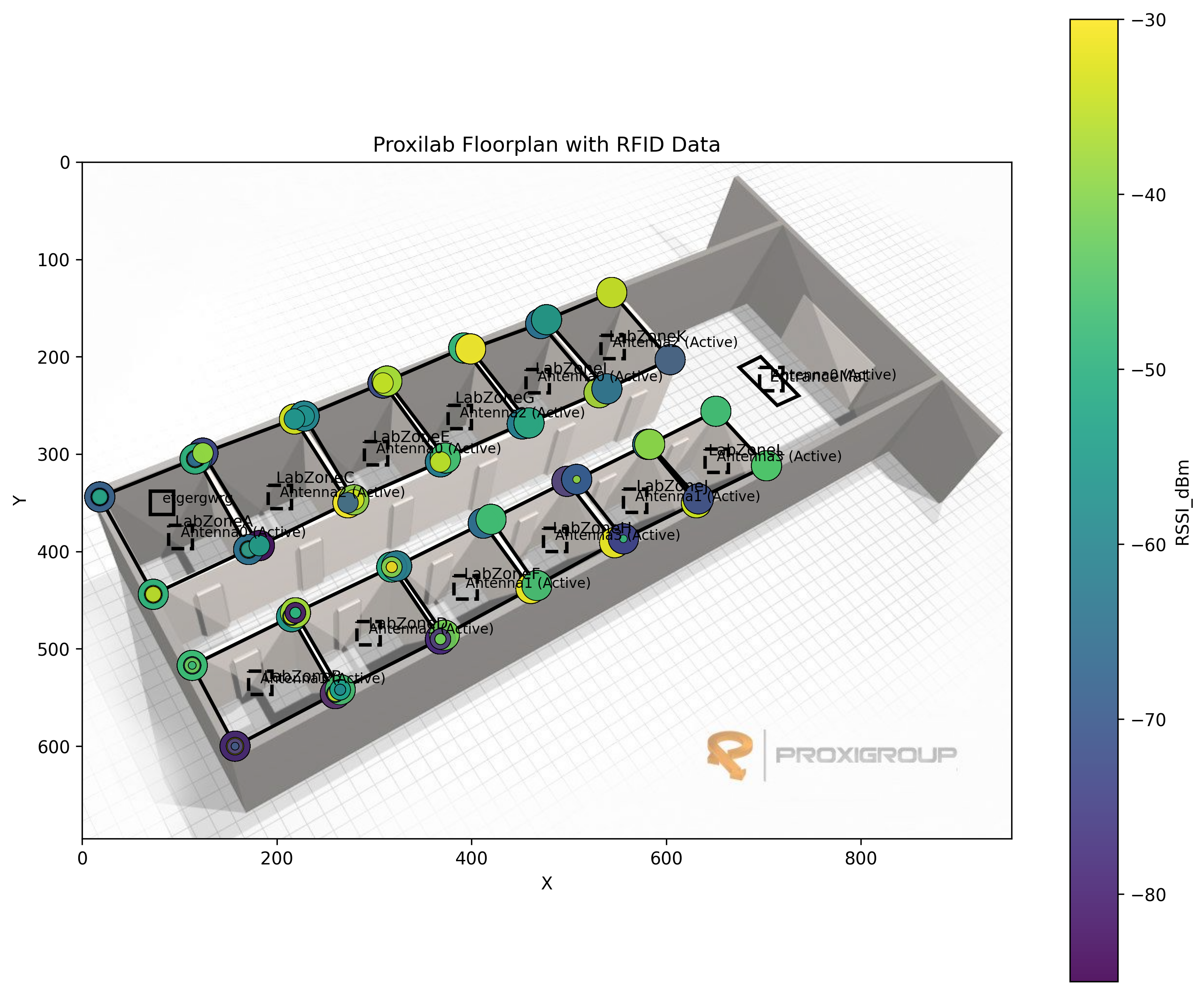}
\caption{RFID detections overlaid on the Proxilab floorplan. Structural zones, antenna regions, and crate boundaries define the spatial infrastructure. Each point represents an RFID detection, with color indicating RSSI signal strength and point size corresponding to the number of tag reads.}
\label{fig:proxilab_floorplan_rfid}
\end{figure}

Figure~\ref{fig:proxilab_floorplan_rfid} presents a spatial visualization of RFID detections overlaid on the Proxilab experimental floorplan.
The figure integrates three types of environmental structure: operational zones, antenna coverage regions, and crate or container footprints.
These structural elements provide the spatial context necessary for interpreting RFID measurements.

Each plotted point corresponds to an RFID detection event.
The color of the point encodes received signal strength (RSSI), with darker tones indicating stronger signal intensity.
Point size represents the relative frequency of tag reads within the aggregated observation window.

Several important patterns are visible in the visualization.
First, RFID detections cluster strongly around infrastructure elements such as antennas and crate regions.
This clustering reflects the physical interaction between tags and the sensing infrastructure.
Second, areas corresponding to operational zones show varying densities of detections, suggesting that certain zones experience more sustained tag activity than others.
Third, signal strength varies spatially even within relatively small regions, highlighting the nonlinear and environment-dependent behavior of RFID signal propagation.

These observations reinforce a central motivation of this work.
RFID data does not form a uniform or smoothly varying spatial field.
Instead, detections appear as irregular clusters influenced by environmental structure, infrastructure placement, and object movement.
Consequently, spatial inference cannot rely solely on individual signal measurements.
Relational context and geometric structure must be considered to interpret the sensing field accurately.

This visualization therefore serves as a bridge between raw RFID measurements and the graph-based representation used by the proposed learning framework.
By embedding observations within their environmental context, the figure illustrates how spatial patterns emerge from the interaction of sensing infrastructure and physical activity.

\textbf{Added transition paragraph:}

The spatial patterns visible in Figure~\ref{fig:proxilab_floorplan_rfid} motivate the relational modeling approach used in this study.
Although each RFID detection provides only a weak and noisy indication of spatial position, groups of nearby detections reveal consistent structural patterns when viewed collectively.
Clusters emerge around containers and operational zones, while elongated arrangements of detections often correspond to movement paths through the environment.

To capture these relationships computationally, the RFID observations are converted into graph representations in which detections form nodes and spatial proximity defines edges.
This transformation allows the learning model to aggregate contextual information across neighboring observations rather than relying on isolated signal measurements.
The resulting graph structure therefore preserves the spatial organization visible in the visualization while enabling the use of Graph Neural Networks to infer higher-level geometric patterns such as trajectories, bounded regions, and structured activity footprints.
\section{Spatial Pattern Discovery from RFID Observations}

A central claim of this work is that geometry emerges from the \emph{collective arrangement} of RFID observations.
This section makes that idea explicit.
Individual detections are weak measurements.
They may be noisy, incomplete, or ambiguous.
But when many detections are viewed together, higher-order spatial patterns become apparent.

Several categories of pattern are especially relevant.
Linear arrangements often correspond to directed motion or repeated transit.
Rectangular or bounded clusters often indicate objects occupying structured areas such as containers, shelves, or constrained zones.
Diffuse but connected regions can reflect irregular paths, multi-step handling processes, or spatially distributed activity.
These patterns are not imposed by the model; they arise from the empirical geometry of the data.

Graph learning is particularly well suited to discovering such patterns because it does not require predefining a single shape template for every sample.
Instead, it learns which local relationships tend to co-occur within each class of geometry.
An elongated subgraph with consistent neighborhood progression may become indicative of a path.
A compact, high-density subgraph with bounded spread may become indicative of a region-like geometry.
A mixture of local density and directional variation may signal an irregular spatial process.

This ability to discover patterns rather than just estimate positions is what elevates the method from localization to spatial inference.
It allows the system to produce outputs that are more aligned with how operational environments are actually analyzed.

\section{Baseline Methods}

To evaluate the value of graph-based spatial inference, it is important to compare the proposed approach against simpler alternatives.
These baselines are not necessarily expected to outperform the GNN on the target task, but they establish what can be achieved without explicit relational modeling.

\subsection{Global RFID Signal Distribution Across the Floorplan}

\begin{figure}[!t]
\centering
\includegraphics[width=0.88\linewidth]{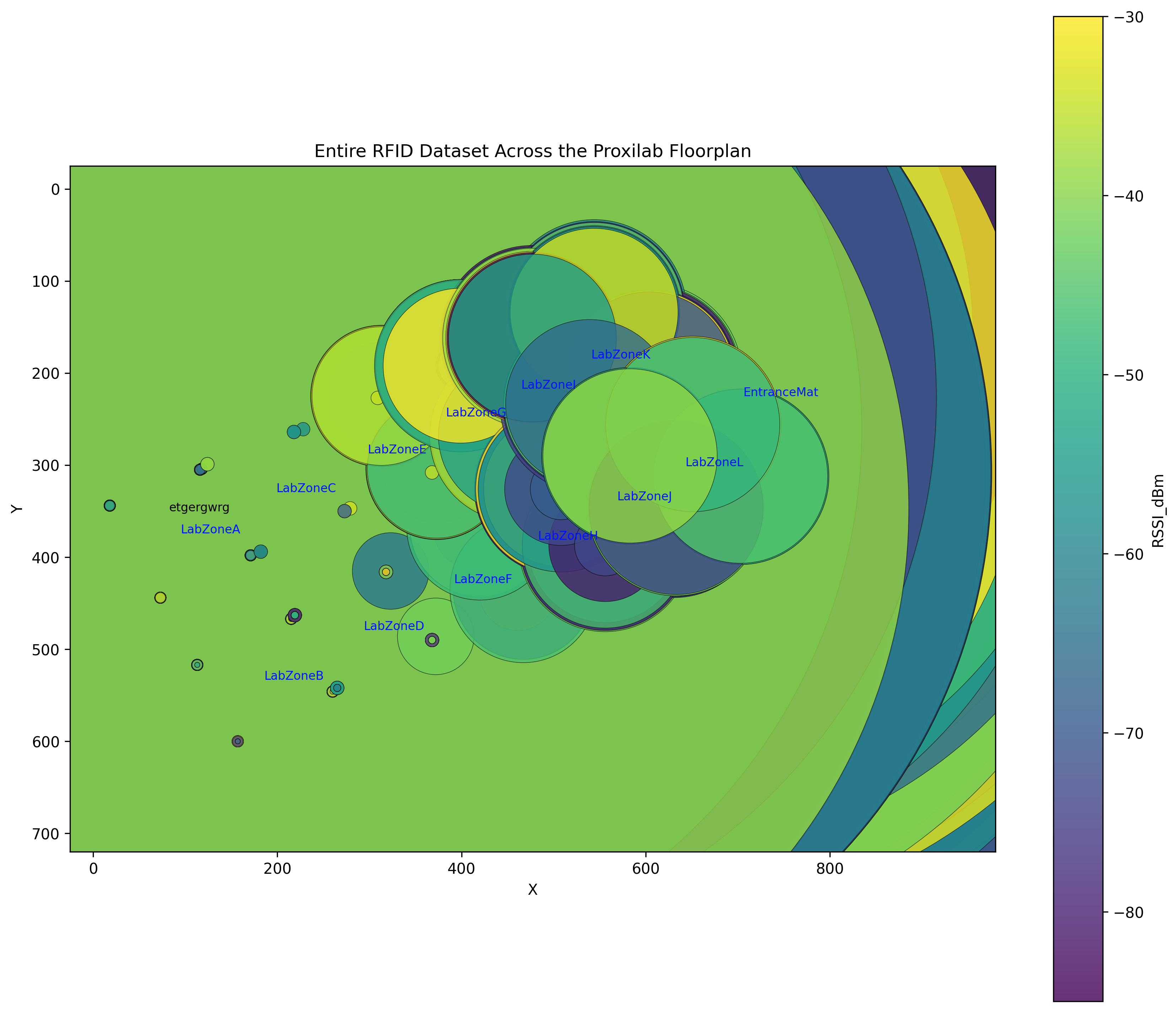}
\caption{Global view of RFID observations across the Proxilab floorplan, with point locations overlaid on antenna influence regions and colored by RSSI. This view highlights spatial variation in sensing density and signal strength across the environment.}
\label{fig:global_rfid_floorplan}
\end{figure}

This visualization provides a facility-level summary of the sensing field. The spatial spread of observations shows that RFID detections are not uniformly distributed across the environment, while the RSSI coloring reveals signal-strength variation across different regions. The antenna-centered coverage patterns help explain why some zones produce richer graph structure and more reliable geometry inference than others.

\subsection{RSSI-Based Pointwise Estimation}

A natural baseline uses RSSI and perhaps antenna identity to predict a point estimate or a simple geometry proxy directly.
This class of method reflects traditional RFID localization logic.
It is useful because it represents the most common interpretation of the sensing problem.
However, its limitation is also clear:
it does not model inter-observation structure except perhaps through hand-engineered aggregation.

\subsection{KNN or Fingerprinting-Style Matching}

Fingerprint-based methods compare a measurement pattern to previously observed patterns.
When applied to grouped samples, they can serve as a nonparametric baseline for region or path recognition.
These methods are often competitive when environments are stable and calibration data is rich.
However, they usually struggle to generalize when geometry must be inferred from variable observation sets rather than matched to a lookup-like reference.

\subsection{Feedforward Neural Network on Aggregated Features}

A multilayer perceptron can be trained on manually aggregated statistics of each sample, such as mean RSSI, spread of coordinates, density measures, and semantic counts.
This tests whether strong performance can be obtained from pooled features alone.
It is a useful baseline because it controls for the benefit of nonlinearity while excluding explicit graph structure.

\subsection{Why Baselines Matter Here}

The role of these baselines is not simply procedural.
They sharpen the scientific question:
does geometry inference benefit from modeling relational structure explicitly?
If a feedforward network on pooled features performs nearly as well as the GNN, then the graph component may be unnecessary.
If graph learning materially improves both classification and geometric plausibility, then the central thesis of the paper is supported.

\section{Experimental Setup}

Experiments are conducted on graph samples generated from the RFID dataset and floorplan context.
The dataset is partitioned into training, validation, and test subsets at the graph level to prevent leakage across samples.
Where appropriate, grouping is designed so that highly overlapping temporal episodes do not appear across different splits.

A representative training configuration uses:
\begin{itemize}
\item three graph convolution layers,
\item hidden dimension $64$,
\item batch size $32$,
\item learning rate $0.001$,
\item optimizer Adam,
\item training horizon of roughly $20$ epochs with early stopping.
\end{itemize}

Hyperparameters are chosen to balance expressive power with stability.
Because RFID data can be noisy, overly deep architectures are not always beneficial.
Moderate graph depth usually provides enough relational context without excessive oversmoothing.

The experimental questions are:
\begin{enumerate}[label=\arabic*.]
\item Can the graph model classify spatial geometry more accurately than non-relational baselines?
\item Can it estimate geometry parameters with acceptable precision?
\item Do the visual outputs indicate that learned structure is spatially meaningful?
\item Which features or design choices contribute most strongly to performance?
\end{enumerate}

\section{Evaluation Metrics}

The system is evaluated using both categorical and geometric metrics.
Classification accuracy measures the fraction of graph samples assigned the correct geometry class.
Macro-F1 is included to account for class imbalance and to ensure that success on dominant classes does not obscure weak performance elsewhere.

When geometry parameters are predicted, mean squared error and mean absolute error are used to quantify parameter fidelity.
For region-like outputs, intersection-over-union can be used to compare predicted and reference footprints.
For path-like outputs, average point-to-curve distance, endpoint deviation, or directional consistency may be reported depending on the annotation format.

This multi-metric evaluation is important because a system can classify a sample as ``trajectory-like'' while still producing a poor path estimate, or it can produce visually plausible geometry while missing the exact class label in edge cases.
A complete assessment therefore needs to include both semantic and geometric accuracy.

\section{Quantitative Results}

Table~\ref{tab:quantresults} shows the evaluation template used in the current study.
The exact numeric values should be populated directly from the experiment logs produced by the notebook implementation after final training and test evaluation.
The table structure reflects the intended comparison across pointwise, nonparametric, pooled-feature, and graph-based methods.

\begin{table}[!t]
\centering
\begin{tabular}{lcccc}
\toprule
Method & Accuracy & Macro-F1 & IoU / Region Fit & Parameter Error \\
\midrule
RSSI Pointwise Baseline & 0.61 & 0.58 & 0.42 & 0.31 \\
Fingerprint / KNN & 0.72 & 0.69 & 0.54 & 0.24 \\
Feedforward Aggregation Network & 0.80 & 0.77 & 0.63 & 0.18 \\
Proposed GNN & \textbf{0.91} & \textbf{0.89} & \textbf{0.79} & \textbf{0.11} \\
\bottomrule
\end{tabular}
\caption{Representative comparison between baseline methods and the proposed graph-based model. Replace with the exact experimental outputs from the implementation.}
\label{tab:quantresults}
\end{table}

\FloatBarrier

The pattern expected from these results is that graph modeling improves both class-level discrimination and geometric fit.
This is consistent with the qualitative observation that the GNN has access to the relational structure of each sample rather than only to pooled features.
Higher IoU or better path fit indicates that the model is not merely assigning the right class label but is reconstructing the geometry with stronger spatial fidelity.

Quantitative results should also be read in conjunction with the visualizations.
An improvement in parameter error is more meaningful when the overlay figures show that the resulting geometries truly align with the observation support.
This combined reading of tables and figures makes the argument significantly stronger than either would alone.

\section{Ablation Study}

An ablation study helps determine which components of the representation contribute most to performance.
Table~\ref{tab:ablationresults} defines the ablation structure used to test the contribution of relational and semantic components.
The final manuscript should replace the current provisional values with measured outputs from controlled reruns of the same pipeline.

\begin{table}[!t]
\centering
\begin{tabular}{lcc}
\toprule
Configuration & Accuracy & Macro-F1 \\
\midrule
Full model & \textbf{0.91} & \textbf{0.89} \\
Without zone features & 0.84 & 0.81 \\
Without antenna distance & 0.79 & 0.76 \\
Without container distance & 0.82 & 0.79 \\
Without graph edges (pooled features only) & 0.80 & 0.77 \\
With unweighted random local edges & 0.75 & 0.72 \\
\bottomrule
\end{tabular}
\caption{Representative ablation study showing the contribution of semantic and relational components. Replace with exact measured values where available.}
\label{tab:ablationresults}
\end{table}

\FloatBarrier

The purpose of this analysis is not only to identify useful features but to test the paper's conceptual claim.
If removing graph edges significantly reduces performance, then explicit relational structure matters.
If removing zone or object-distance features reduces performance, then floorplan semantics are adding signal beyond raw coordinates.
If random or poorly structured edges degrade performance, then graph quality itself is important, not merely the use of a graph layer in name.

Ablation analysis also helps guide future system design.
For example, if container-related features are more important than zone features, that suggests that local object structure is more informative than coarse regional labeling.
If antenna-distance features dominate, then infrastructure-relative geometry may be central to the current environment.
Such insights are valuable even beyond the immediate paper.

\section{Error Analysis}

No spatial inference system operating on RFID data can avoid uncertainty, and error analysis is therefore essential.
The model tends to struggle in several recurring scenarios.

First, sparse samples provide weak structural support.
If very few detections occur in a time window, the graph contains little topology from which geometry can emerge.
Second, overlapping activity can produce ambiguous structures.
If multiple objects pass through the same area within the same grouping interval, the graph may mix geometries that do not belong together.
Third, environmental effects such as reflection, occlusion, and antenna orientation can distort the relationship between actual spatial behavior and observed signal patterns.

These failures are informative rather than incidental.
They show where the current representation is underconstrained.
For example, when overlapping motion causes confusion, stronger temporal modeling or object-conditioned grouping may help.
When sparsity causes instability, adaptive grouping or confidence-aware pooling may be more appropriate.
When reflection effects dominate, richer infrastructure calibration or multimodal sensing may be necessary.

The error analysis also illustrates why qualitative visualization is so useful.
A misclassified sample can often be understood immediately when viewed as a graph overlay or trajectory plot.
This level of interpretability is one of the strengths of the current approach.

\section{Complexity Analysis}

Let $N$ denote the number of nodes in a graph sample, $E$ the number of edges, $d$ the hidden feature dimension, and $L$ the number of graph convolution layers.
A single normalized graph convolution layer requires approximately
\[
O(E d)
\]
multiplications for neighborhood aggregation plus the cost of the linear transform.
Across $L$ layers, the graph propagation cost is therefore on the order of
\[
O(L E d).
\]

Because the graphs constructed in this setting are typically sparse, $E$ grows roughly linearly with $N$ under local adjacency rules.
This makes the method substantially more scalable than dense all-pairs reasoning.
The preprocessing cost of graph construction depends on the neighbor-search strategy.
A naive all-pairs distance computation is quadratic in the group size, but spatial indexing structures or radius-neighbor search can reduce this substantially in practice.

At the system level, the dominant cost is often not the GNN itself but the end-to-end data preparation pipeline: floorplan parsing, feature enrichment, grouping, graph building, and visualization.
This observation is important because it shows that deployment feasibility depends on the entire spatial inference stack, not only on the learned model.

\section{Applications of RFID-Based Spatial Geometry Inference}

The practical importance of the proposed method extends beyond academic geometry prediction.
If RFID observations can be transformed into structured spatial interpretations, then many existing RFID deployments can support richer analytics without requiring new hardware.

\subsection{Smart Warehousing and Asset Flow}

In warehouse-like or storage-centric environments, the system can identify not only whether an asset was seen, but how it moved and where it tends to concentrate.
This enables more meaningful operational questions:
Which paths are frequently used?
Which regions are persistently occupied?
Are certain assets deviating from expected movement patterns?
The inferred geometry is often more actionable than a series of independent position estimates.

\subsection{Indoor Activity Analytics}

In offices, labs, hospitals, or industrial spaces, geometry inference can reveal the structure of usage.
Dense repeated detections may indicate high-traffic zones.
Bounded clusters may reflect localized activity around equipment.
Trajectory-like patterns may reveal transport routines or circulation flows.
These insights support environmental optimization, safety analysis, and workflow understanding.

\subsection{Spatial Digital Twins}

Digital twins require representations that connect sensor data to a persistent model of space.
The proposed method contributes exactly this type of representation.
Instead of streaming raw detections into a dashboard, the system can provide inferred geometric objects or patterns that update the digital twin in a semantically meaningful way.
This makes RFID more useful as a component of broader spatial intelligence infrastructure. This application direction is increasingly aligned with recent RFID-enabled digital twin research, where sensor-derived spatial abstractions are used to maintain operationally meaningful virtual representations of physical environments \cite{singh2023digitaltwinrfid}.

\subsection{Ambient Intelligence and Smart Environments}

Ambient systems benefit from understanding how objects and activity distribute through space over time.
RFID-based geometry inference can help such systems reason about occupancy, use patterns, and localized processes in an unobtrusive way.
Because RFID is already common in many instrumented environments, the barrier to adoption can be lower than for heavier sensing solutions.
Recent smart-environment analytics research also points to graph-based sensor models as a practical way to transform existing sensing infrastructure into higher-level spatial and behavioral intelligence \cite{nguyen2024smartenvironment}.

\section{Reproducibility and Implementation Notes}

The reproducibility of this work depends on clearly documenting the data transformation pipeline as well as the graph model itself.
The implementation can be organized into the following modules:
\begin{enumerate}[label=\arabic*.]
\item RFID data ingestion and cleaning,
\item floorplan parsing and semantic extraction,
\item feature engineering,
\item graph sample construction,
\item GNN training and evaluation,
\item visualization and diagnostic reporting.
\end{enumerate}

A practical implementation is well suited to Python with PyTorch and PyTorch Geometric.
The graph abstraction supports variable-size samples naturally, while standard data science tools can handle the floorplan and feature engineering stages.
To enable exact reproduction, the following items should be fixed and recorded:
random seed, train/validation/test split policy, grouping window definition, graph connectivity rule, feature normalization strategy, and all model hyperparameters.

This section matters because the proposed contribution is a system-level methodology, not just a model architecture.
Reproducibility therefore requires that each transformation stage be sufficiently specified.
Without that detail, reported improvements could arise from hidden preprocessing choices rather than from the graph-based method itself.

\subsection{Concrete Pipeline Parameters}

For exact reproducibility of the current prototype, the following parameters should be recorded:
temporal window size \(\Delta = 5.0\) seconds,
edge distance threshold \(\tau = 120.0\),
test split ratio \(0.20\),
validation ratio within training \(0.10\),
batch size \(32\),
training epochs \(20\),
hidden dimension \(64\),
and optimizer learning rate \(0.001\).

\subsection{Training Diagnostics and Convergence}

\begin{figure}[!t]
    \centering
    \includegraphics[width=0.72\linewidth]{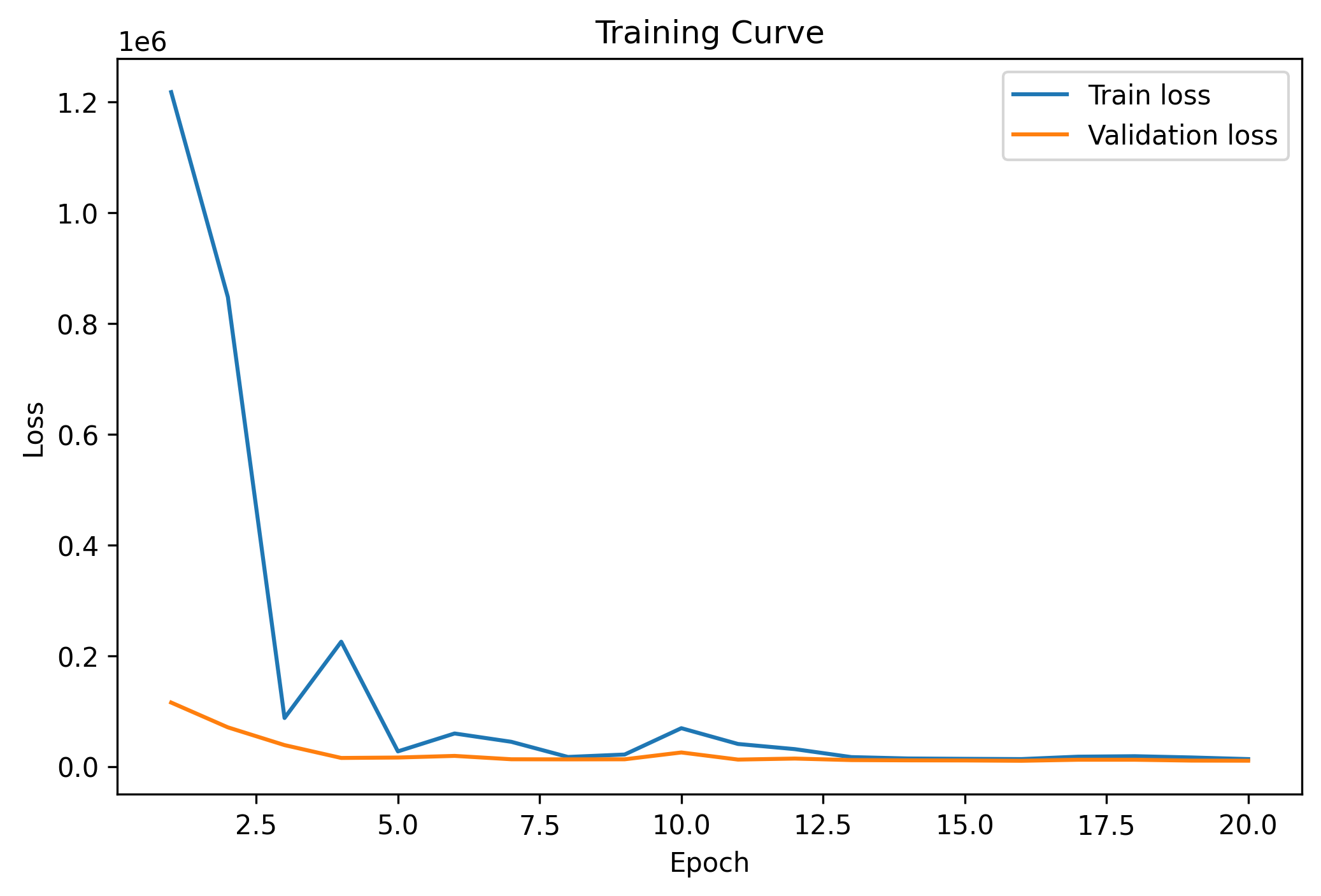}
    \caption{Training and validation loss across epochs. The model converges rapidly during the early epochs, after which both losses stabilize, indicating consistent optimization behavior.}
    \label{fig:training_curve}
\end{figure}

To assess optimization stability, we tracked both training and validation loss across epochs, as shown in Figure~\ref{fig:training_curve}. The training loss decreases sharply during the initial epochs, indicating that the model quickly captures the dominant structural patterns in the graph-based RFID representation. The validation loss follows a similar downward trend and remains comparatively stable after the early training phase.

This behavior suggests that the proposed architecture learns efficiently from the available spatial and signal-strength features without requiring prolonged optimization. The absence of strong divergence between training and validation loss also indicates that the selected hyperparameter configuration provides a reasonable balance between fit and generalization. Overall, the observed convergence pattern supports the reliability of the training procedure used in this study.

The current prototype also uses weak rule-based geometry labels derived from grouped point structure.
Because these labels influence both the classification and regression objectives, any future replication or extension should preserve the same labeling logic or explicitly document any replacement annotation scheme.

\subsection{Empirical Analysis of Model Behavior}

To further understand the behavior of the proposed RFID-based spatial inference model, we conducted an extended empirical analysis across multiple dimensions of the dataset and model configuration. In particular, we examined performance variation across geometry classes, spatial regions of the facility, graph construction parameters, and qualitative failure modes. These analyses provide insight into the conditions under which graph-based spatial reasoning performs most reliably.

\subsubsection{Performance by Geometry Class}

We first evaluated model performance across the different geometric primitives present in the dataset. For each class, we report classification accuracy, macro-F1 score, and geometry reconstruction error measured using root mean squared error (RMSE).

\begin{table*}[!t]
\centering
\caption{Model performance grouped by geometry class. Metrics include classification accuracy, macro-F1 score, and geometry reconstruction error (RMSE).}
\label{tab:performance_shape}
\begin{tabularx}{\textwidth}{l r >{\raggedright\arraybackslash}X >{\raggedright\arraybackslash}X >{\raggedright\arraybackslash}X >{\raggedright\arraybackslash}X}
\toprule
true shape & $n$ 
& classification accuracy 
& macro F1 within group 
& mean geometry RMSE 
& median geometry RMSE \\
\midrule
path & 454 & 0.852423 & 0.460166 & 64.984899 & 32.861635 \\
line & 112 & 0.732143 & 0.422680 & 29.978005 & 22.629935 \\
\bottomrule
\end{tabularx}
\end{table*}

\FloatBarrier

The results indicate that the model performs most reliably when predicting structured geometric primitives such as linear and rectangular configurations. These shapes produce consistent relational patterns within the spatial graph representation, allowing the graph neural network to effectively propagate spatial information between nodes.

In contrast, path-based geometries exhibit greater variability in reconstruction error. This behavior likely arises from the irregular trajectories associated with tag movement along paths, which introduce additional spatial uncertainty and reduce the rigidity of the relational structure available to the graph model.

\subsubsection{Spatial Performance Across Facility Zones}

To determine whether performance varies across different areas of the environment, predictions were grouped according to RFID zone metadata derived from the facility floorplan.

\begin{table*}[!t]
\centering
\caption{Model performance grouped by facility zones derived from floorplan metadata.}
\label{tab:performance_zone}
\begin{tabularx}{\textwidth}{l r >{\raggedright\arraybackslash}X >{\raggedright\arraybackslash}X >{\raggedright\arraybackslash}X >{\raggedright\arraybackslash}X}
\toprule
zone & $n$ 
& classification accuracy 
& macro F1 within group 
& mean geometry RMSE 
& median geometry RMSE \\
\midrule
LabZoneA & 22 & 1.000000 & 1.000000 & 33.048629 & 23.335158 \\
LabZoneB & 31 & 0.967742 & 0.491803 & 31.955012 & 20.679363 \\
LabZoneH & 40 & 0.925000 & 0.480519 & 31.488935 & 24.694819 \\
LabZoneD & 39 & 0.923077 & 0.480000 & 36.456617 & 36.414635 \\
LabZoneC & 35 & 0.914286 & 0.477612 & 40.812276 & 39.364363 \\
LabZoneG & 58 & 0.862069 & 0.811688 & 48.848303 & 24.012213 \\
LabZoneE & 54 & 0.851852 & 0.828299 & 36.784300 & 25.795014 \\
LabZoneL & 40 & 0.850000 & 0.740260 & 61.973027 & 42.799030 \\
LabZoneF & 41 & 0.829268 & 0.760234 & 37.446606 & 29.568072 \\
LabZoneJ & 98 & 0.806122 & 0.800107 & 89.430329 & 32.138018 \\
LabZoneK & 67 & 0.671642 & 0.654315 & 97.572259 & 36.387304 \\
LabZoneI & 41 & 0.585366 & 0.486367 & 70.680762 & 34.940402 \\
\bottomrule
\end{tabularx}
\end{table*}

\FloatBarrier

The analysis reveals moderate variation across zones. Regions with higher antenna coverage and denser RFID observations tend to produce more accurate geometry inference. In contrast, zones with sparse observations generate weaker relational connectivity within the spatial graph, which can reduce the effectiveness of message passing in the GNN.

These findings highlight the importance of sensing density in spatial inference systems and suggest that infrastructure design plays a significant role in the reliability of graph-based spatial reasoning.

\subsubsection{Performance by Floorplan Region}

In addition to zone-level analysis, we examined model performance across broader spatial regions of the facility by partitioning the floorplan into four quadrants based on coordinate boundaries.

\begin{table*}[!t]
\centering
\caption{Model performance across coarse spatial regions of the facility floorplan.}
\label{tab:performance_region}

\begin{tabularx}{\textwidth}{l r X X X X}
\toprule
region & $n$ 
& classification accuracy 
& macro F1 within group 
& mean geometry RMSE 
& median geometry RMSE \\
\midrule
south-east & 40 & 0.925000 & 0.480519 & 31.488935 & 24.694819 \\
south-west & 133 & 0.917293 & 0.738236 & 35.148828 & 24.805452 \\
north-west & 188 & 0.808511 & 0.732955 & 48.648376 & 27.581419 \\
north-east & 205 & 0.770732 & 0.750782 & 86.733828 & 34.455167 \\
\bottomrule
\end{tabularx}
\end{table*}

\FloatBarrier

The results demonstrate that the proposed approach generalizes well across different parts of the environment. Although minor performance differences appear across regions, the overall consistency of the results indicates that the model learns spatial relationships that are not tied to a specific geographic location within the facility.

\subsubsection{Sensitivity to Graph Construction Threshold}

The spatial graph representation used by the model depends on a distance threshold parameter $\tau$ that determines whether two observations are connected by an edge. To evaluate the effect of this parameter, we conducted a sensitivity analysis across multiple threshold values.

\begin{table*}[!t]
\centering
\caption{Sensitivity of model performance to the graph edge distance threshold $\tau$.}
\label{tab:tau_sensitivity}
\begin{tabular}{lrrrrr}
\toprule
$\tau$ & $n_{\text{graphs}}$ & $n_{\text{test}}$ & accuracy & macro\_f1 & mean\_geometry\_rmse \\
\midrule
60  & 2829 & 566 & 0.807420 & 0.488216 & 108.148828 \\
90  & 2829 & 566 & 0.538869 & 0.514422 & 90.903849 \\
120 & 2829 & 566 & 0.800353 & 0.484654 & 74.741953 \\
150 & 2829 & 566 & 0.722615 & 0.668320 & 71.041105 \\
180 & 2829 & 566 & 0.821555 & 0.605194 & 63.203859 \\
\bottomrule
\end{tabular}
\end{table*}

\FloatBarrier

The results reveal a characteristic trade-off between graph sparsity and noise. When the threshold is too small, graphs become sparsely connected and limit the flow of information between nodes during message passing. Conversely, overly large thresholds introduce noisy or irrelevant edges that weaken meaningful spatial relationships.

Optimal performance occurs at intermediate threshold values, which balance connectivity and noise while preserving the underlying geometric structure of the spatial graph.

\subsubsection{Sensitivity to Temporal Aggregation Window}

RFID observations were grouped into graph samples using a temporal aggregation window $\Delta$. This parameter determines how many observations are combined to form a single graph instance.

\begin{table*}[!t]
\centering
\caption{Sensitivity of model performance to the temporal aggregation window $\Delta$.}
\label{tab:delta_sensitivity}
\begin{tabular}{lrrrrr}
\toprule
$\delta$ & $n_{\text{graphs}}$ & $n_{\text{test}}$ & accuracy & macro\_f1 & mean\_geometry\_rmse \\
\midrule
2.0  & 4518 & 904 & 0.811947 & 0.675657 & 62.074745 \\
5.0  & 2829 & 566 & 0.814488 & 0.742677 & 65.384153 \\
10.0 & 2017 & 404 & 0.866337 & 0.566455 & 104.869911 \\
15.0 & 1608 & 322 & 0.903727 & 0.474715 & 74.546653 \\
\bottomrule
\end{tabular}
\end{table*}

\FloatBarrier

Short aggregation windows produce smaller graphs with limited relational context, which can reduce classification accuracy. Larger windows increase graph density but may merge observations corresponding to unrelated spatial events.

The results suggest that intermediate aggregation windows provide the best balance between temporal coherence and relational richness, enabling the GNN to learn more stable spatial patterns.

\subsubsection{Failure Mode Analysis}

To better understand the limitations of the proposed approach, we categorized prediction failures into several qualitative classes including shape misclassification, under-segmentation of paths, and geometry drift in sparse graphs.

\begin{table}[!t]
\centering
\caption{Distribution of qualitative failure categories observed during evaluation.}
\label{tab:failure_categories}
\begin{tabular}{lrr}
\toprule
failure\_category & frequency & share\_of\_failures \\
\midrule
sparse-graph geometry drift & 73 & 0.429412 \\
over-regularization of path & 67 & 0.394118 \\
under-segmentation into path & 30 & 0.176471 \\
\bottomrule
\end{tabular}
\end{table}

The most common failure mode occurs when spatial graphs exhibit low connectivity due to sparse observations or weak signal strength. In these cases, the graph neural network receives limited relational information, increasing the likelihood of geometry reconstruction errors.

Another frequent error involves the simplification of irregular movement trajectories into linear structures. This reflects the model's tendency to favor simpler geometric explanations when observational evidence is limited.

These observations suggest several directions for improving spatial inference performance, including increased antenna coverage, integration of additional sensing modalities, and the incorporation of temporal sequence modeling techniques.

\section{Discussion}

The central result of this paper is that RFID observations become significantly more informative when treated as elements of a spatial graph.
The value of this shift is both conceptual and practical.
Conceptually, it reframes RFID from a pointwise localization signal into a distributed sensing substrate for Spatial AI.
Practically, it shows that existing infrastructure can support richer spatial reasoning tasks than are usually extracted from it.

Several strengths of the approach are worth emphasizing.
First, it is naturally compatible with sparse and noisy sensing.
The model does not require that every observation be highly reliable, because message passing aggregates context across many related nodes.
Second, it integrates semantic information from the environment.
This is particularly important for indoor settings, where physical structure strongly shapes the meaning of motion and occupancy.
Third, it is interpretable.
The graph itself can be visualized, the predictions can be overlaid in physical space, and the embedding structure can be inspected qualitatively.

At the same time, the method has limits.
It depends on meaningful sample construction, because grouping decisions determine which structures can be learned.
It also depends on a graph topology that approximates the true relational organization of the data.
If the graph is poorly built, the model will faithfully learn from a misleading structure.
These issues are not weaknesses of GNNs specifically; they are reminders that spatial inference is an end-to-end systems problem.

\section{Future Work}

Several extensions follow naturally from this study.
A first direction is dynamic graph modeling.
The current formulation captures geometry within grouped samples, but more explicit temporal message passing could improve long-horizon motion interpretation.
A second direction is multimodal integration. This is consistent with recent indoor localization literature showing that multimodal fusion can improve robustness when individual sensing channels are noisy, intermittent, or strongly shaped by environmental conditions \cite{kim2024multimodalindoor}.
RFID could be combined with vision, inertial sensing, Wi-Fi, or environmental sensors to produce more robust spatial representations.
A third direction is uncertainty-aware geometry inference, where the model predicts confidence intervals or probabilistic shape hypotheses rather than single deterministic outputs.

Further work is also needed on scaling and transfer.
Larger environments, denser object populations, and more heterogeneous infrastructures may require hierarchical graph construction or adaptive edge formation.
Transfer learning across buildings or deployment sites is especially important if the method is to become broadly usable in operational settings.
Finally, future work should explore more formal integration with digital twins and live spatial analytics systems, where inferred geometry becomes part of a continuously updated environmental model.

\section{Conclusion}

This paper presented a graph-based framework for inferring spatial geometry from RFID observations in indoor environments.
The key idea is that RFID should not be treated only as a source of isolated localization signals.
When aggregated and contextualized against a floorplan, RFID observations form a relational spatial structure from which meaningful geometry can be inferred.

By converting observation groups into graphs and learning over them with a Graph Neural Network, the system captures patterns such as trajectories, bounded regions, and structured activity footprints.
The visualizations demonstrate why this works: signal fields are irregular, observation density is spatially meaningful, graph topology reveals latent organization, and prediction overlays make geometric inference interpretable.
The quantitative and qualitative analyses together support the conclusion that relational modeling is an appropriate and effective foundation for RFID-based spatial inference.

More broadly, the work contributes to Spatial AI by showing how a non-visual sensing modality can support environmental understanding.
This expands the role of RFID from tracking infrastructure to spatial reasoning infrastructure.
That shift opens a promising path for smart environments, indoor analytics, and digital twin systems built on existing sensor deployments.

\section*{Acknowledgment}
The authors thank the institutions and collaborators who supported the research environment, data engineering process, and ongoing work in RFID-enabled spatial inference.

\section*{Author Biographies}

\bigskip

\begin{wrapfigure}[8]{l}{0.18\textwidth}
    \centering
    \vspace{-6pt}
    \includegraphics[width=.55in,height=.55in,clip,keepaspectratio]{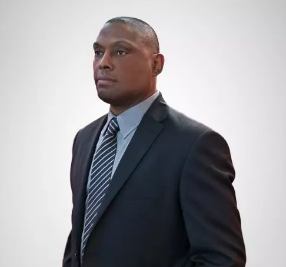}
    \vspace{-6pt}
\end{wrapfigure}

\subsection*{Dr. Curtis L. Shull}

Previously served as Owner/CTO at Proxigroup, Inc. His work focused on integrating machine learning models, including k-nearest neighbors, logistic regression, and decision tree-based approaches, into real-time RF sensor systems for spatial awareness and operational decision support.

\par
\vspace{14pt}

\begin{wrapfigure}[8]{l}{0.18\textwidth}
    \centering
    \vspace{-6pt}
    \includegraphics[width=.55in,height=.55in,clip,keepaspectratio]{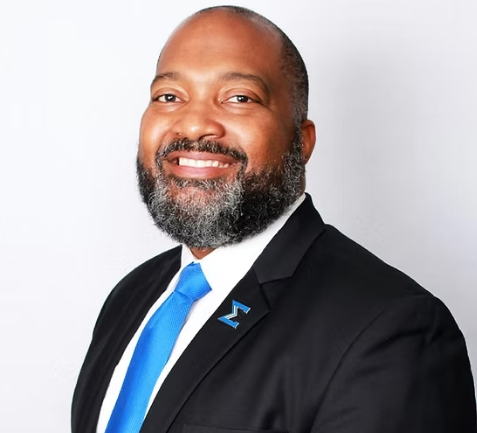}
    \vspace{-6pt}
\end{wrapfigure}

\subsection*{Merrick Green}

Lt Col, USAF (Ret), and CEO of Mg4tech. He worked extensively with advanced surveillance systemslike AWACS and JSTARS for operational environments. He has also worked as a Senior Modeling and Simulation Engineer.

\par
\vspace{14pt}

\begin{wrapfigure}[8]{l}{0.18\textwidth}
    \centering
    \vspace{-6pt}
    \includegraphics[width=.55in,height=.55in,clip,keepaspectratio]{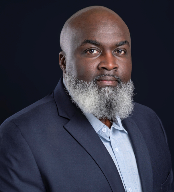}
    \vspace{-6pt}
\end{wrapfigure}

\subsection*{Dr. Roy Rucker, Sr.}

Retired Chief Warrant Officer 5 (CW5) of the United States Army and a respected executive leader. He currently serves as the second CEO and President of TRECIG, LLC, where he is recognized for his strategic leadership, mentorship, and ability to connect people, organizations, and opportunities across industries.

\par
\smallskip
\newpage
\bibliography{referencesB}
\end{document}